%% file: main.tex
\documentclass{article}
\pdfoutput=1

\usepackage{lineno,hyperref}
\usepackage[utf8]{inputenc}
\usepackage{multirow}
\usepackage[htt]{hyphenat}
\usepackage{bbm}
\usepackage{makecell}
\usepackage{amsmath}
\usepackage{amssymb}
\usepackage{graphicx}
\usepackage{booktabs}
\usepackage{caption}
\usepackage{subcaption}
 
\graphicspath{{figs/}}

\title{A Temporal Convolutional Network-Based Approach and a Benchmark Dataset for Colonoscopy Video Temporal Segmentation}
\author{
    Carlo Biffi$^{1}$\thanks{Corresponding author: cbiffi@cosmoimd.com}, 
    Giorgio Roffo$^{1}$, 
    Pietro Salvagnini$^{1}$, 
    Andrea Cherubini$^{1,2}$
}
\date{
    $^{1}$Cosmo Intelligent Medical Devices, Dublin, Ireland\\
    $^{2}$Milan Center for Neuroscience, University of Milano–Bicocca, Milano, Italy
}

\begin{document}

\maketitle

\begin{abstract}
\textbf{Background and Objective:} Following recent advancements in computer-aided detection and diagnosis systems for colonoscopy, the automated reporting of colonoscopy procedures is set to further revolutionize clinical practice. A crucial yet underexplored aspect in the development of these systems is the creation of computer vision models capable of autonomously segmenting full-procedure colonoscopy videos into anatomical sections and procedural phases. In this work, we aim to create the first open-access dataset for this task and propose a state-of-the-art approach, benchmarked against competitive models.

\textbf{Methods:} 
We annotated the publicly available REAL-Colon dataset, consisting of 2.7 million frames from 60 complete colonoscopy videos, with frame-level labels for anatomical locations and colonoscopy phases across nine categories. We then present ColonTCN, a learning-based architecture that employs custom temporal convolutional blocks designed to efficiently capture long temporal dependencies for the temporal segmentation of colonoscopy videos. We also propose a dual k-fold cross-validation evaluation protocol for this benchmark, which includes model assessment on unseen, multi-center data.

\textbf{Results:} 
ColonTCN achieves state-of-the-art performance in classification accuracy while maintaining a low parameter count when evaluated using the two proposed k-fold cross-validation settings, outperforming competitive models. We report ablation studies to provide insights into the challenges of this task and highlight the benefits of the custom temporal convolutional blocks, which enhance learning and improve model efficiency.

\textbf{Conclusions:} 
We believe that the proposed open-access benchmark and the ColonTCN approach represent a significant advancement in the temporal segmentation of colonoscopy procedures, fostering further open-access research to address this clinical need. Code and data are available at: \url{https://github.com/cosmoimd/temporal_segmentation}.
\end{abstract}

\section{Introduction}
Optical colonoscopy is crucial for early detection and removal of premalignant polyps (adenomas) to prevent colorectal cancer (CRC), the third most common cancer and second in mortality \cite{dekker_advances_2018, zorzi2023adenoma}. This procedure is divided into two main phases: intubation (or insertion), during which the endoscope is navigated to the ceacum (noted as the cecal intubation or insertion time), and withdrawal. The withdrawal phase involves a thorough examination of the colon mucosa looking for polyps (withdrawal time) and concludes once the endoscope is fully retracted from the anus. Recent years have seen the emergence of the first commercial computer-aided detection (CADe) and diagnosis (CADx) systems, promising to significantly enhance adenoma detection rates (ADR), which measures the proportion of screening colonoscopies in which one or more adenomas are identified, and to help standardizing colonoscopy procedures \cite{berzin_position_2020, spadaccini_computer-aided_2021, biffi_novel_2022}. 

Concurrent with worldwide adoption of colonoscopy as the gold standard screening tool for CRC, international medical societies have advocated for the monitoring of key colonoscopy quality metrics related to interval CRC rates \cite{shine2020quality,rex2023key}.  These quality indicators include the ADR alongside the withdrawal time (minimal of 6 minutes, target time of 10 minutes), cecal and ileum intubation rates, and the Boston Bowel Preparation Scale (BBPS) score. The BBPS score assesses colon cleanliness by grading the visibility of the intestinal mucosa and it is derived from the sum of scores assigned to three segments of the colon: the right colon, the transverse colon, and the left colon \cite{rex2023key}. CAD(e/x) systems capable of automatically extracting these quality metrics from full-procedure video data at the procedure's conclusion, incorporating into automated reports these metrics alongside a precise summary of polyp count, optical diagnosis, and anatomical location, could provide a standardized means to facilitate quality control in day-to-day clinical practice, conduct large-scale endoscopic skills assessment evaluations, and monitor progress during colonoscopy training, ultimately reducing colonoscopy times and costs while improving screening effectiveness \cite{tavanapong2022artificial,gimeno2023artificial}.  

Several computer-based methods that enable automated measurement of a number of metrics that reflect the quality of the colonoscopic procedure have been recently introduced \cite{hwang2005automatic,liu2010automated,katzir2022estimating,chang2022development,de2023automated}. Hwang et al. \cite{hwang2005automatic} proposed to estimate colonoscopy timings, such as withdrawal time, by analyzing camera motion. Subsequent studies have aimed to estimate withdrawal time either through direct frame-by-frame caecum detection \cite{de2023automated, chang2022development, liu2010automated} or by integrating model outputs for egomotion, depth, and classification \cite{katzir2022estimating}.  Furthermore, research has targeted the automatic computation of the BBPS score using frame or short-video clips classification \cite{zhou2021multi,feng2023development}. Refining this application, however, necessitates accurate segmentation of video frames depicting the ascending, transverse, and descending colon segments to enhance BBPS score calculation. Furthermore, temporally segmenting additional anatomical structures, such as the sigmoid or rectum, would be important for improving automated polyp reporting, as locations like the recto-sigmoid can influence surveillance intervals. In the same fashion, the definition of the caecum and ileum segments could enhance algorithms aimed at detecting caecum and ileum intubation rates. 

From a computer vision standpoint, addressing the above-mentioned clinical needs requires the ability to analyze and segment full-procedure colonoscopy videos into semantically meaningful sections. This task parallels challenges encountered in temporal video segmentation and automated surgical workflow analysis, where considerable effort has been invested in segmenting videos into distinct actions, surgical steps, or phases \cite{ding2023temporal,demir2023deep}. Early approaches in these fields leveraged image classification networks for frame-by-frame action or phase identification, which later evolved to incorporate Recurrent Neural Networks (RNNs) to better capture temporal dependencies \cite{jin2017sv,zisimopoulos2018deepphase}. However, RNNs struggle with long sequence processing, prompting a shift towards Temporal Convolutional Networks (TCNs), particularly encoder-decoder and multi-layered TCNs with deformable or dilated convolutions \cite{lea2017temporal,lei2018temporal}, which have demonstrated superior performance \cite{ding2023temporal}. The introduction of the Multi-Stage TCN (MS-TCN) \cite{farha2019ms,li2020ms}, with its multi-stage architecture and truncated MSE loss, as seen in the TeCNO approach for surgical phase recognition \cite{czempiel2020tecno}, further enhanced video sequence segmentation accuracy through iterative refinements. More recent developments have focused on Transformers \cite{yi2021asformer,ramesh2021multi,czempiel2021opera,gao2021trans,ding2022exploring}, yet these models face challenges when processing long videos due to the high computational cost of attention mechanisms across extended sequences. ASFormer \cite{yi2021asformer}, which employs a hierarchical model to progressively capture local-to-global temporal relations through stacked dilated convolutions and windowed self-attention, was introduced to mitigate these computational demands \cite{ding2023temporal}. Nonetheless, ASFormer's computational overhead remains substantial for processing long videos, such as colonoscopy recordings that often exceed an hour in duration. Moreover, effective training of Transformer-based models generally requires large datasets, with the statistical unit in the case of video temporal segmentation being the number of videos. This is crucial for learning a diversity of spatio-temporal patterns and transitions. However, typical video datasets provide only hundreds, rather than millions, of videos. All these limitations have led to the development of specialized architectures tailored for the specific task to ensure successful deployment \cite{yi2021asformer,czempiel2021opera, gao2021trans,demir2023deep}.

While a recent effort has evaluated MS-TCN approaches and ASFormer for labeling colonoscopy video segments across various categories \cite{kelner2023semantic}, their application and assessment on full-procedure video data remain unexplored. Full-procedure colonoscopy videos pose unique challenges, including their extended duration, a large number of frames with poor visibility and segments such as the ceacum or ileum that may vary in duration or be absent in some videos. Addressing these challenges requires methods capable of effectively capturing the global context from lengthy video sequences lasting 15 minutes to 1 hour or more, thus requiring large temporal receptive fields, while discerning patterns across varying temporal scales from seconds to minutes. TCN-based approaches typically require stacking a series of dilated 1D convolutions to exponentially increase the model's receptive field \cite{kelner2023semantic, czempiel2020tecno, demir2023deep}. This study aims to pioneer the task of temporally segmenting full-length colonoscopy videos by proposing ColonTCN, a TCN-based approach consisting of custom temporal convolutional blocks that exploit double-dilated residual convolutions and regularization. These blocks are designed to better learn from video data in this novel application, while also resulting in more compact models.

Research in temporal action segmentation and surgical workflow analysis has significantly advanced thanks to publicly available datasets such as GTEA, 50Salads, and Breakfast for the former, and M2CAI16, Cholec80, and Cataract-101 for the latter \cite{ding2023temporal,demir2023deep}. However, the collection of full-procedure colonoscopy videos poses challenges similar to those encountered in surgical video collection, including patient data protection, hardware/recording limitations, and storage constraints. We believe that the lack of a publicly available annotated dataset has notably hindered research progress and model benchmarking in this domain. To address this gap, we introduce annotations for the 60 full-resolution, unaltered colonoscopy videos for the open-access REAL-Colon dataset \cite{biffi2024real}, enabling researchers to openly benchmark approaches in this new domain. Specifically, we release annotations for three colonoscopy phases (outside the colon, insertion, and withdrawal phase) and seven distinct colon segments (ceacum, ileum, ascending colon, transverse colon, descending colon, sigmoid, and rectum).  These annotations facilitate the computation of colonoscopy timings, such as withdrawal time, enable algorithm benchmarking for colon segment recognition crucial for polyp diagnosis (e.g., rectum-sigmoid colon), and allow for the computation of quality metrics, such as ceacum and ileum intubation rates and computing the BBPS score (ascending, transverse, and descending colon).

In summary, this work makes the following contributions:

\begin{itemize}
\item We have annotated the 2.7 million frames of the REAL-Colon dataset \cite{biffi2024real}, comprising 60 full-procedure videos from six centers for colonoscopy video temporal segmentation, enabling open-access model benchmarking.
\item We propose a novel 9-class video classification task leveraging the REAL-Colon dataset's labels and a model evaluation framework. We introduce metrics for model performance assessment and a dual-validation model evaluation strategy within a k-fold cross-validation framework, including model assessment on unseen, multi-center data.
\item We propose ColonTCN, a TCN-based approach using custom temporal convolution blocks, tailored for temporally segmenting colonoscopy videos. We compare it against competitive approaches, achieving state-of-the-art results in both classification accuracy and parameter efficiency.
\item We offer comprehensive ablation studies to assess the influence of various ColonTCN model components, loss functions, and data augmentation techniques, while also shedding light on the challenges of this novel task.
\end{itemize}

We believe these contributions provide a significant advancement in automated colonoscopy video analysis, offering valuable resources for the community to address this new problem while establishing baselines and a state-of-the-art approach. The remainder of this paper is organized as follows: Section \ref{sec:met} details the annotations released for the REAL-Colon dataset, describes the proposed ColonTCN architecture for colonoscopy video understanding and outlines the proposed model evaluation framework. Section \ref{sec:res} details implementation details together with reporting models results and ablation studies. Finally, Section \ref{sec:dis} discusses the results of this work and future directions.

\section{Methods}
\label{sec:met}
This section outlines the annotation process and characteristics of the frame-level annotations released in this work for the REAL-Colon dataset (Section \ref{subsec:realcolon}). Then, it offers a comprehensive explanation of TCN and MS-TCN models utilized for the temporal segmentation of colonoscopy videos (Section \ref{subsec:tcn}) and of proposed TCN architecture, ColonTCN, and provides a description of the adopted loss functions for its optimization (Section \ref{subsec:loss}). Finally, Section \ref{subsec:dataset} introduced the training and evaluation protocol that we propose to benchmark models on the REAL-Colon dataset for the temporal video segmentation task.

\subsection{Annotation of the REAL-Colon dataset}
\label{subsec:realcolon}
The REAL-Colon dataset \cite{biffi2024real} is an open-access dataset comprising 2.7 million high-resolution frames extracted from 60 full-procedure colonoscopy videos, capturing a diverse patient population across four cohorts from six medical centers spanning three continents. "The first two dataset cohorts, Cohort 1 and Cohort 2, include 15 videos each from clinical studies NCT03954548 and NCT04884581 on ClinicalTrials.gov, acquired across three U.S. centers and one Italian center, respectively. Cohorts 3 and 4 also consist of 15 videos each, sourced from separate acquisition campaigns at centers in Austria and Japan \cite{biffi2024real}. The video recordings document each procedure from its initiation near the entrance of the colon to its conclusion as the endoscope is withdrawn. For comprehensive details on the REAL-Colon data acquisition, dataset construction, patient statistics, and privacy considerations, we refer readers to \cite{biffi2024real}.

\input{fig_text/fig1a}
\input{fig_text/fig1b}

For this study, all 60 videos in the REAL-Colon dataset underwent additional annotation. Eight medical image annotation specialists, supervised by an expert gastroenterologist from a pool of three experts, followed a standardized process using a specialized in-house annotation tool capable of timestamp placement. The process included multiple feedback rounds and can be summarized as follows. First, three colonoscopy phase labels — outside, insertion, and withdrawal — were annotated, each with a start and stop frame. Instruction was to make the withdrawal phase start when the endoscope reached the appendiceal orifice or ileocecal valve, and the endoscopist initiated cecum exploration following cleaning. The transition between the colon interior and exterior was marked at the frame where the anus was entered or exited. Then, the withdrawal phase was annotated into seven colon segment sub-classes: ceacum, ileum, ascending colon, transverse colon, descending colon, sigmoid colon, and rectum. The ascending, transverse, and descending colon were delineated by identifying the first frame in which the hepatic and splenic flexure was visible. Similarly, the cecum was defined by identifying the ileocecal valve and appendiceal orifice, while the rectum was defined by identifying the rectosigmoid junction. In cases where the endoscope entered the ileocecal valve and explored the terminal ileum, distinguished by its specific mucosal appearance, the corresponding label was assigned to those frames.

As a result, at the end of the annotation process, only a small percentage of frames (0.2\%) were classified as uncertain, all coming from only two videos. These frames correspond to instances of brief recording interruptions or cases where an expert gastroenterologist could not confidently assign a frame label during frame class transitions. Figure \ref{fig:fig1a} illustrates the sequence of segments and the proportional distribution of each label class within the dataset resulting from the annotation process. Annotations have been released here: \url{https://doi.org/10.6084/m9.figshare.26472913}. 

Figure \ref{fig:fig1a} reveals the pronounced class imbalance typical of colonoscopy procedures, with the outside and ileum class frames being significantly less frequent in the dataset (in 34 out of 60 videos the ileum was not reached). Figure \ref{fig:fig1b} complements this by presenting a box and whisker plot depicting the percentage of video length occupied by each frame label in the REAL-Colon dataset, highlighting the substantial intra-class and inter-class variability inherent to the problem.

\subsection{ColonTCN}
\label{subsec:tcn}
Each colonoscopy video \( \mathbf{V} \in \mathbb{R}^{T \times H \times W \times 3} \) consists of \( T \) frames, each with dimensions \( H \times W \) and 3 color channels. TCN approaches typically involve two key steps:

\begin{enumerate}
    \item A visual frame feature extraction step, where video frames are embedded into a low-dimensional representation \( \mathbf{X} \in \mathbb{R}^{T \times D} \). In line with several previous works on temporal video segmentation and surgical workflow analysis, in this work we utilize ResNet-50 pretrained on ImageNet as the feature extractor \cite{he2016deep}, yielding frame representations of size \( D=2048 \).
    \item A TCN model that performs spatio-temporal reasoning across the entire set of video frames \( \mathbf{X} \), outputting predictions \( \mathbf{Y} \in \mathbb{R}^{T \times C} \), where \( C \) represents the number of distinct frame labels to be predicted. Each prediction for the \( i \)-th frame is influenced by both preceding and succeeding frames in the context of  colonoscopy video temporal segmentation since the analysis is conducted using acasual temporal convolutions at the procedure end.
\end{enumerate}

In this work, we propose ColonTCN for this second step. Similarly to TCN-based approaches adopted in similar fields \cite{czempiel2020tecno}, ColonTCN commences with an initial feature reduction (FR) layer comprising a \(1 \times 1\) convolutional layer paired with a ReLU activation function (Figure \ref{fig:fig2a}). This layer is designed to align the dimensions of the input features \( \mathbf{X} \) obtained with the ResNet-50 model with the feature map $\mathbf{H} \in \mathbb{R}^{T \times F}$ input of the first temporal block (TB) of ColonTCN, where $F \ll D$ enables the construction of shallower models as illustrated in Figure \ref{fig:fig2b}. Following this, the network includes a series of \(l\) TBs applied to the features maps \( \mathbf{H}_{l-1} \) of the previous layer. Each TB of the ColonTCN architecture, specifically designed for this task, contains two weight-normalized, dilated, acausal 1D convolutional layers with kernels \(\mathbf{W}_{1,l}\) and \(\mathbf{W}_{2,l}\), biases \(\mathbf{b}_{1,l}\) and \(\mathbf{b}_{2,l}\) and dilation factor \(2^l\). Acasual convolutions are utilized to ensure that the prediction for the \( i \)-th frame is informed by both preceding and succeeding frames. After each convolutional layer, a Rectified Linear Unit (ReLU) activation function and dropout regularization are applied, as described in the following equations:

\begin{equation}
\mathbf{C}_{1,l} = \text{DropOut}[\text{ReLU}(\mathbf{W}_{1,l} * \mathbf{H}_{l-1} + \mathbf{b}_{1,l})]
\end{equation}

\begin{equation}
\mathbf{C}_{2,l} = \text{DropOut}[\text{ReLU}(\mathbf{W}_{2,l} * \mathbf{C}_{1,l} + \mathbf{b}_{2,l})]
\end{equation}

Additionally, each block integrates a residual connection to facilitate gradient flow and enable training of deeper networks:

\begin{equation}
\mathbf{H}_l = \text{ReLU}(\mathbf{H}_{l-1} + \mathbf{C}_{2,l})
\end{equation}

The addition of each TB allows the models' receptive field to expand exponentially. The use of double dilated convolutions in these blocks further broadens the models receptive field, enhancing the capture of global temporal dependencies. The use of dropout, weight regularization and residual connections enables better gradient flow even when stacking several of these TBs. In contrast, approaches such as TeCNO \cite{czempiel2020tecno} instead only stack dilated convolutions, which we have found to lead to sub-optimal learning when trying to reaching large receptive fields and lead to the proposal of this TB architecture. Finally, the output class probabilities are computed by applying a \(1 \times 1\) convolution to the last dilated convolution layer's output, followed by a softmax activation function: $ \mathbf{Y} = \text{Softmax}(\text{Conv}_{1 \times 1}(\mathbf{H}_l)) $. This results in the output prediction matrix \( \mathbf{Y} \in \mathbb{R}^{T \times C} \) for the video \( \mathbf{V}\).

In a MS-TCN model, such as the ones proposed in TeCNO \cite{czempiel2020tecno}, the output \( \mathbf{Y}^1 \) from the initial TCN stage is refined through \( M \) additional TCN stages. In this work, we apply the same approach to ColonTCN, yielding MS-ColonTCN. Specifically, in a multi-stage model each stage processes the output from its preceding stage and, while all stages adhere to the architecture of the first stage, they possess distinct weight sets. The only difference in subsequent stages is the input feature size set to \( C \), eliminating the need for the FR layer. Instead, tensors are directly input to the first TB of each stage. These stages sequentially output a series of tensors \( \mathbf{Y}^S \) for \( S=1 \) to \( M \), representing the class probabilities for each stage.

\input{fig_text/fig2a}
\input{fig_text/fig2b}

\subsection{Loss Function}
\label{subsec:loss}
In the training of TCN models for surgical workflow analysis, a classification loss is typically computed at each frame $t$ using weighted cross-entropy loss:

\begin{align}
L_{\text{cls}} = \frac{1}{T} \sum_{t} - w_c \; y_{t,c} \; \log(\hat{y}_{t,c})
\end{align}

where class weights \( w_c \) are determined using median frequency balancing to address class imbalances \cite{czempiel2020tecno}, and \( \hat{y}_{t,c} \) represents the predicted probability for the ground truth label \( c \) at time \( t \).

Multi-stage architectures typically also employ truncated mean squared error over the frame-wise log-probabilities to reduce over-segmentation errors \cite{li2020ms}

\begin{align}
L_{\text{T-MSE}} = \frac{1}{TC} \sum_{t,c} \tilde{\Delta}_{t,c}^2
\end{align}
\begin{equation}
\tilde{\Delta}_{t,c} = 
\begin{cases} 
\Delta_{t,c} & \text{if } |\log y_{t,c} - \log y_{t-1,c}| \leq \tau, \\
\tau & \text{otherwise},
\end{cases}
\end{equation}

The overall loss, computed over the  \( M \) stages, thus becomes

\begin{align}
L &= \frac{1}{M} \sum_{M} L_{\text{cls}} + \lambda L_{\text{T-MSE}} 
\end{align}

where \( \lambda \) is a balancing weight. In this work, this same loss function was adopted for the training of ColonTCN and competitive approaches using $\tau=4$ and $\lambda=0.15$.

\subsection{Model Evaluation Framework}
\label{subsec:dataset}

We propose both a 5-fold and a 4-fold cross-validation (CV) method to train and evaluate models on the 60 full-procedure videos of the REAL-Colon dataset we have annotated. For the 5-fold CV approach, we randomly distributed the dataset into subsets comprising 44 training videos, 4 validation videos, and 12 testing videos. This division was achieved by randomly selecting videos from the four clinical cohorts within the REAL-Colon dataset, ensuring that each cohort contributed 11, 1, and 3 videos to the training, validation, and testing subsets, respectively. This sampling approach ensured the uniqueness of the videos in each of the five testing sets, allowing for a thorough evaluation of the models across the entirety of the REAL-Colon dataset. Random selection was repeated multiple times to make certain that each test fold included at least one video depicting the ileum, sourced from a minimum of three out of the four cohorts. The specifics of the selected video distribution across the 5-fold splits are detailed in the Supplementary Materials. The 4-fold CV method was instead designed to include two studies from the REAL-Colon dataset in the training set, one in the validation set, and one in the testing set. This CV experiment aimed to test the models' ability to generalize to unseen centers and in a more challenging scenario (less training data too). Indeed, each model in this setting is trained on only 30 videos, as opposed to the 44 used in the 5-fold scenario.

\input{tables/folds}

Consistent with established practices in the field, we standardized the frame rate of all videos to 5 frames per second (fps) for both training and inference \cite{czempiel2020tecno,demir2023deep}. The length of these videos showed considerable variation, with an average of 8,471 frames (28 minutes), a standard deviation of 4,947 frames (16 minutes), and a range from a minimum of 3,485 (11 minutes) to a maximum of 24,772 frames (82 minutes) per video. Table \ref{tbl:folds} provides a comprehensive breakdown of the frame class distributions in the test sets from both the 5-fold and 4-fold CV evaluations of the REAL-Colon dataset, with videos adjusted to 5 fps. It is noteworthy that Cohort 3 includes longer videos than the other studies and none of the videos in Cohort 2 reached the ileum, presenting an additional real-world challenge for the 4-fold CV evaluation.

In both cross-validation (CV) settings, performance evaluation for a given model is computed for all $C=9$ classes by computing the average F1 score per-frame, as commonly done in surgical workflow analysis research \cite{demir2023deep}. Using the set of frames labeled as Ground Truth (GT) for a class \(c\) and the set of frame Predictions (P) predicted for that class, the F1 score represents the harmonic mean of precision and recall over all the frames labeled to that class:
\[
\text{Precision (PR)} = \frac{|GT \cap P|}{|P|}, \quad \text{Recall (RE)} = \frac{|GT \cap P|}{|GT|}, \quad \text{F1} = \frac{2 \cdot PR \cdot RE}{PR + RE}.
\]

Moreover, for each model, three additional metrics are computed, split by split, over all videos of the REAL-Colon dataset:

\begin{itemize} 
    \item The class-weighted average F1 score (wF1) accounts for the inherent class imbalances within the dataset, offering a more nuanced understanding of the models' performance by assigning greater importance to less frequent classes. The weighted F1 score is defined as:
    \[
    \text{wF1} = \sum_{c=1}^{C} w_c \cdot \text{F1}_c,
    \]
    where \(w_c\) is the weight for class \(c\) and \(\text{F1}_c\) is the F1 score for class \(c\).
    
    \item The weighted Jaccard index (wJacc). The Jaccard index is defined as:
    \[
    \text{Jaccard Index (J)} = \frac{|GT \cap P|}{|GT \cup P|}.
    \]
    and measures the models' temporal segmentation accuracy by evaluating the overlap between predicted and actual segments. Its weighted version is:
    \[
    \text{wJacc} = \sum_{i=1}^{C} w_c \cdot \text{J}_c,
    \]
    
    \item The Mean Absolute Percentage Error (MAPE) metric for the withdrawal time (WMAPE). The withdrawal time is defined as the time spent (derived by counting the frames) neither outside the colon nor during the insertion phase. Accurate withdrawal time estimation relies on the precise segmentation of short sequences (e.g., frames outside the colon) and a comprehensive understanding of the long temporal sequences to identify the start of the withdrawal phase. The WMAPE measures the deviation between the predicted and actual number of withdrawal frames, employing the absolute mean percentage error to account for the varying lengths of withdrawal times present in the dataset:
    \[
    \text{WMAPE} = \frac{1}{N} \sum_{i=1}^{N} \left| \frac{A_i - P_i}{A_i} \right| \times 100,
    \]
    where \(A_i\) is the actual number of withdrawal frames, \(P_i\) is the predicted number of withdrawal frames and N is the number of videos.
\end{itemize}

Finally, we profile each model to estimate its computational efficiency by reporting the number of parameters and Giga Floating-Point Operations (GFLOPs), which quantify the floating-point operations required for a forward pass through the model.

\section{Results}
\label{sec:res}

\subsection{Implementation details}
\label{subsec:idetails}
Feature extraction was conducted using a ResNet-50 model, pretrained on the ImageNet dataset, with video input frames resized to a 224x224 resolution before feature extraction\cite{he2016deep}. For enhancing the model's robustness, two types of data augmentation were applied to all videos: spatial and temporal. Spatial augmentation included a random combination of flipping, mirroring, and slight color adjustments in the HSV color space applied consistently across all video frames, subsequently encoded using ResNet-50. Temporal augmentation involved encoding videos at their original frame rates, followed by a random temporal subsampling in half of the instances, or subsampling one over the subsampling factor of the total frame count in the other instances. Unless specified otherwise, we applied both spatial and temporal augmentations with a 50\% probability to each video during the training of each model. Once sampled, the same augmentation was applied to all the frames of a video. Frames annotated as 'uncertain' (constituting less than 0.2\% of the dataset frames and all belonging to two videos) were removed.

For model training, the best optimizer in our experiments was AdamW utilized with an initial learning rate of 5* $10^{-4}$, a weight decay factor of $10^{-2}$ and a linear learning rate scheduler to gradually reduce the learning rate to $10^{-6}$ over the training period.  This optimization scheme was used for all models as well as the values of the hyperparameters $\tau=4$ and $\lambda=0.15$ in the loss function, as this configuration provided the best results following a preliminary grid search. All models underwent training for 30,000 iterations in the 5-fold CV setting and 22,000 iterations in the 4-fold CV setting, as these durations provided the best results. However, ASFormer was trained for only 10,000 iterations, as shorter training yielded better outcomes for this model. The optimal model for each fold being selected based on the highest wF1 score on the validation dataset after 5,000 iterations, addressing occasional peaks, happening to all models, in the validation set that did not yield good results on the test set. All model training was conducted using a batch size of 6 on an NVIDIA A100 GPU, with all code developed in PyTorch. Batching was implemented by padding the feature representations of shorter videos to match the length of the longest video in each batch and then selectively computing the losses by applying a mask to ignore the padded frames. All models discussed in this study feature an output Softmax layer, with the frame prediction being the class with the highest score. 

\input{tables/main_tab}

\subsection{Comparison with the Baselines}
Table \ref{tbl:maintab} presents a detailed comparison between the ColonTCN model proposed in this study, which leverages stacks of TB, and other competitive approaches. ColonTCN incorporates double weight-normalized dilated residual 1x1 convolutions with a kernel size of 7 and 64 feature channels in each TB, further optimized with dropout regularization, as shown in Figure \ref{fig:fig2b}. For the 5-fold and 4-fold scenarios, 13 and 12 TBs were employed, respectively.

In Table \ref{tbl:maintab}, we also report the results for the TCN and MS-TCN models  as introduced in the TeCNO paper using acausal convolutions, both of which yielded the best results on both benchmarks with 14 levels \cite{czempiel2020tecno}\footnote{https://github.com/tobiascz/TeCNO}. Additionally, results of the transformer-based approach by ASFormer \cite{yi2021asformer}\footnote{https://github.com/ChinaYi/ASFormer/} are included. ASFormer features 10 blocks and 64 features—the most extensive configuration supportable by an NVIDIA A100 GPU. For this model, we noticed significant training instability; thus, we report only the best results over four training sessions.

For both the 5-fold and 4-fold settings, Table \ref{tbl:maintab} includes the F1 score for each class, the wF1, the wJacc, and the WMAPE computed on all 60 dataset videos, along with the number of parameters and GFLOPs for each model to demonstrate their computational complexity. The ResNet50 model used as a feature extractor for all the models has a footprint of 25.6M params resulting in 4.185 GFLOPs. Overall, the 4-fold CV experiments proved to be more challenging, with the ileum class emerging as the most difficult, alongside the delineation of the descending colon. The outside colon class is the easiest since it offers semantically different frames compared to all other frames in the dataset, which are all colon frames. Notably, the WMAPE is significantly affected by the drop in segmentation performance for the insertion and outside colon classes in the 4-fold CV scenario.

ColonTCN achieved the highest wF1 and wJacc scores and the lowest WMAPE, while competing with ASFormer, which employs more parameters, for the best F1 score on individual classes. It can also be noticed how ColonTCN uses half the parameters of MS-TCN and still yields a better result. These results underscore the superior performance and greater parameter efficiency of ColonTCN compared to ASFormer and TeCNO models. In the Supplementary Materials, we present the recomputation of Table \ref{tbl:maintab} without employing class weighting through median frequency balancing. This yielded the same model rankings but produced lower results, due to the models' difficulties in learning from less represented classes.
ColonTCN achieved the highest wF1 and wJacc scores and the lowest WMAPE, while competing with ASFormer, which employs more parameters, for the best F1 score on individual classes. It can also be noticed how ColonTCN uses half the parameters of MS-TCN and still yields a better result. These results underscore the superior performance and greater parameter efficiency of ColonTCN compared to ASFormer and TeCNO models. In the Supplementary Materials, we present the recomputation of Table \ref{tbl:maintab} without employing class weighting through median frequency balancing. This yielded the same model rankings but produced lower results, due to the models' difficulties in learning from less represented classes.

\input{tables/besttcn}
\input{tables/levels}

\input{tables/stages}
\subsection{ColonTCN Ablation Study}
In Table \ref{tbl:besttcn}, we present an ablation study that underscores the key architecture contributions within ColonTCN that lead to the best models reported in Table \ref{tbl:maintab}. A comparison between the first and second rows demonstrates the effectiveness of the feature reduction (FR) layer over a top-down strategy of gradually decreasing feature dimensions from 512 to 64 within the TBs, which could have been an alternative option. This FR layer utilizes a 1x1 convolution followed by a non-linear activation to reduce the dimensionality of ResNet50 features from 2048 to 64 before they are processed by the TBs, as shown in Figure \ref{fig:fig2a}. Moreover, the integration of residual dilated convolutions (highlighted in the 'Residual' column) and the use of double dilated 1x1 convolutions within each TB (indicated in the 'Double' column), as opposed to single dilated 1x1 convolutions as used in TeCNO, significantly boosted model performance. This enhancement is the key reason for the superior performance of the ColonTCN approach compared to the competitive TeCNO approach. Additionally, incorporating dropout as a regularization method in each TB, specifically a dropout rate of 0.5 following each dilated 1x1 convolution as shown in Figure \ref{fig:fig2b} (noted in the 'DropOut' column), further enhanced performance. All these results favour the construction of the TBs as proposed in our ColonTCN approach. 

Table \ref{tbl:levels} presents an ablation study on the number of TBs used in ColonTCN. Selecting 11 blocks achieves a temporal receptive field of 24,565 frames, equivalent to 80 minutes, suitable for almost every colonoscopy video. However, interestingly in our experiments using 13 blocks in the 5-fold CV scenario and 12 in the 4-fold CV scenario proved more effective, probably leading to more refined high-level temporal features for this task. The variation between the models can be attributed to the 5-fold model operating on more data and encountering a less challenging test environment, where having one more TB likely improves learning.  

\input{tables/abl_aug}

\input{tables/losses}
\input{fig_text/fig3}

\subsection{MS-ColonTCN Models}
Table \ref{tbl:stages} presents results for several MS-ColonTCN architectures studied for this using varying levels (number of TBs in each ColonTCN) and number of refinement stages in on the REAL-Colon dataset. The obtained results reveal that, in both the 4-fold and 5-fold CV settings, multi-stage architectures do not outperform single-stage architectures without refinement (0 stages, reported in the first row of each table). Notably, the introduction of one refinement stage leads to a significant drop in performance in both settings, which is only partially mitigated by adding more stages. This effect is particularly evident when differentiating among the more challenging classes, such as the ileum, ascending, transverse, and descending colon segments. This challenge likely arises from the inherent difficulty in modeling these classes in the refinement stages based solely on their temporal sequences at the input. Of note, the best MS-ColonTCN models of Table \ref{tbl:stages} both outperformed single-stage ColonTCN models with 11 levels of Table \ref{tbl:levels} in both CV settings. However, by adding more TB in ColonTCN instead of multiple stages produces better performing and more memory efficient models.

\subsection{ColonTCN Model Optimization}
Table \ref{tbl:abl_aug} provides detailed insights into the impact of temporal and spatial augmentations on the best ColonTCN model. The impact of weighted Cross-Entropy (wCE) and Temporal Mean Squared Error (TMSE) losses is instead depicted in the ablation studies reported in Table \ref{tbl:losses}. Collectively, these augmentations and loss functions contribute to improved performance across all classes, with a particularly notable enhancement observed in the ileum, the most challenging class. Conversely, experiments with weighted Focal loss \cite{lin2017focal}, exploited in the attempt to further improve performance on difficult classes, did not yield performance improvements, as reported in Table \ref{tbl:losses}.

\subsection{Visual Results}
Finally, Figure \ref{fig:fig3} showcases visual representations of class predictions made by the best ColonTCN and MS-ColonTCN models on six random videos of the REAL-Colon dataset, compared to those generated by ASFormer, the best competitive model from Table \ref{tbl:maintab}. Models were obtained from the 5-fold CV experiments. It can be observed that ASFormer, due to its shorter temporal receptive field, struggles to achieve a global understanding of the entire video predicting the rectum during insertion (third column, second row). In contrast, when comparing the MS-TCN and TCN models, we noticed that the MS-TCN model tends to segment videos in larger clusters of frames with the same prediction (for example, the rectum class in the second column, second row video), struggling to obtain fine-grained predictions. 

\section{Discussion}
\label{sec:dis}

In this study, our objective was to establish the foundation for developing automated, learning-based temporal segmentation models for full-procedure colonoscopy videos, aiming to enhance automated reporting at the procedure conclusion.
By introducing the first extensive collection of frame-wise annotations for the REAL-Colon dataset \cite{biffi2024real}, we have filled a substantial void in the publicly available resources, providing the first public benchmark for model development and evaluation. This initiative brings the field of colonoscopy video analysis into alignment with related areas such as temporal video segmentation and surgical workflow analysis, which have significantly benefited from open-source initiatives \cite{ding2023temporal,demir2023deep}. Subsequently, we investigated the novel problem of simultaneously segmenting colonoscopy phases and anatomical locations, framing it as a 9-class classification challenge. Additionally, we developed an evaluation protocol comprising two k-fold cross-validation settings to rigorously benchmark the proposed models: a 5-fold configuration optimized to maximize training data, and a more challenging 4-fold configuration designed to assess performance on unseen data and requiring models to train on fewer videos.

As the visualization of the GT annotation for the videos in Figure \ref{fig:fig3} shows, effective temporal segmentation requires models capable of segmenting videos into clusters of same-class frames that can span several minutes (insertion phase) or just a handful of seconds (ileum phase). Additionally, almost all classes exhibit significant inter-class duration variability (Figure \ref{fig:fig1b}); for instance, the search for polyps can prolong navigation for several minutes in the colon segments, and classes such as the ceacum or the ileum may or may not appear. While both the colonoscopy phases and the colon segments follow an ordinal succession, models should also be able to achieve fine-grained, non-causal predictions, such as accurately modeling the end of the insertion phase, detecting the potential presence of the ileum and cecum classes, and capturing the possible back-and-forth transitions between these structures and the ascending colon. Taken together, all these data characteristics make the problem of temporal segmentation of full-procedure videos a challenging one. 

A key contribution of this research was the effective adaptation of TCN-based models to address this problem. In particular, we propose ColonTCN, a specialized TCN architecture characterized by multiple custom TBs forming a deep hierarchy. These TBs, through the exploitation of double dilated residual convolutions and regularization, enable frame-by-frame contextual understanding while efficiently achieving a sufficient temporal receptive field to comprehend the entire procedure.  Two categories of models were compared against ColonTCN: ASFormer and TCNs and MS-TCNs from TeCNO \cite{yi2021asformer, czempiel2020tecno}. The latter were chosen as the most advanced TCN-based architectures in surgical workflow analysis, which do not rely on domain-specific modifications. The former was selected because it is a modern, lightweight transformer-based model and was used in similar research \cite{kelner2023semantic}. From the results reported in Table \ref{tbl:maintab} it can be seen how both class of methods struggle to achieve an adequate receptive field without significantly increasing their total number of parameters. In the case of ASFormer, despite using windowed attention, it is computationally expensive to model such long videos due to the inherently high computational cost of attention mechanisms, which scale quadratically with the length of the input sequence. In the case of models from TeCNO, these models require more temporal convolution layers than ColonTCN to achieve a comprehensive receptive field, which can hinder model learning by adversely affecting gradient flow. The TBs proposed for ColonTCN have shown to better and more efficiently learn from colonoscopy video data.

Between the 5-fold and 4-fold cross-validation settings, the latter proved more challenging as expected, a difference that can be attributed to multiple factors. Firstly, REAL-Colon cohort data were acquired at different centers, resulting in variability in endoscope models and brands which impacts the colon image acquisition process, along with variations in endoscopist skills and performance (notably, no cecum intubation was performed in Cohort 002). By training on data from a centre and testing on a different one, generalization gaps arise for all models. Additionally, the 4-fold setting trains models on only 30 videos compared to 44 in the 5-fold setting, creating a data-scarce environment that further challenges learning. Due to these issues, we believe 4-fold split is highly relevant as it allows us to benchmark which model better learns and generalizes with less data.

Our findings reveal that multi-stage refinements (Table \ref{tbl:stages}), typically beneficial in other domains \cite{farha2019ms,li2020ms,czempiel2020tecno}, did not confer the same advantage over a ColonTCN model equipped with more TBs than necessary to cover the receptive field of the longest video in the dataset (Table \ref{tbl:levels}). We attribute this to the inherent difficulty of modeling colon classes during the refinement stages based solely on their input temporal sequences. Indeed, models with only a single refinement stage experience a significant drop in performance (Table \ref{tbl:stages}, Stages=1 ), while stacking more TBs in a ColonTCN model (instead of refinement stages) results in better performance. Our proposed augmentation strategies significantly enhanced model accuracy, particularly for less represented classes, as shown in Table \ref{tbl:abl_aug}. This improvement was also supported by the use of class-weighted cross-entropy (CE) loss with median frequency balancing, as detailed in the Supplementary Materials. Future work could explore further data augmentation strategies, such as in-painting techniques to artificially manipulate videos by adding or substituting segments from other videos and alternative class-balanced loss functions. Additionally, employing smoother loss functions near transitions could enhance delineation of regions, as demonstrated in surgical workflow analysis \cite{demir2023deep}.

We believe that the contributions of this work could significantly enhance the accuracy of automated colonoscopy reporting systems. Our findings demonstrate that segmenting entire colonoscopy videos in a single pass is feasible, and this capability holds promise for critical applications such as precise estimation of withdrawal times and cecal intubation rates, both key quality indicators in colonoscopy that impact colorectal cancer detection rates \cite{shine2020quality,rex2023key}. Being able to solve the 9-class problem proposed with this work would also enable tracking of procedure durations and accurate identification of colon segments for BBPS scoring (colon into right, center, and left segments—corresponding to the ascending, transverse and descending, rectum, and sigmoid labels in the released dataset). These annotations also support polyp detection reporting, where coupling the information predicted by a ColonTCN model with bounding box detection and polyp characterization data could detailed reporting of each detected polyp. In this study, we report a Mean Absolute Percentage Error (MAPE) of 3.1\% on withdrawal time for the best model in the 5-fold CV setting, which increases to 16.8\% when testing the model on unseen center data in the more challenging 4-fold CV setting (Table \ref{tbl:maintab}). While these results are a promising start, there is potential for improvement in model generalization to unseen centers, and the proposed annotations and k-fold splits facilitate research in this direction.

Our ColonTCN approach, with its offline processing capability—except for frame encoding, which can readily be performed in real-time—can be seamlessly integrated into existing clinical workflows providing feedback at the end of the procedure. Integration would require adding a start-and-stop recording mechanism, either manually via manual intervention or automatically through a model detecting procedure start and end, which is achievable given that transitions into and out of the colon are easily detectable, as our results have shown. These findings also point to potential real-time applications. Although our current models use acausal convolutions, adapting ColonTCN for real-time use with causal temporal convolutions and more compact models is a promising direction for future research. Potential applications include real-time colon location recognition to enhance the optical diagnosis of polyps, verifying key procedural milestones like cecum and ileum intubation and real-time quantification of the duration since the procedure and withdrawal began.

This study does face limitations. While we made efforts to annotate the 2.7 million frames of the REAL-Colon dataset, we believe that there is scope for dataset expansion, as the 60 videos of the REAL-Colon dataset, despite being from multiple centers, still represent a limited population. We are optimistic that our efforts will inspire the creation of additional datasets or the formation of community challenges to promote even more extensive model benchmarking and validation. A limitation of our ColonTCN approach is its limited efficiency in integrating global and local temporal information. Future work could enhance this by adopting a hierarchical segmentation strategy: initially segmenting procedures at a coarse scale, for example working at 1 frame every second, and subsequently applying finer segmentation for detailed analysis. Additionally, the feature extraction method employed, based on a ResNet50 model pretrained on natural images, is not optimized for colonoscopy data. We believe that advancements could be achieved through domain-specific pre-training strategies and the adoption of more sophisticated encoders as done in similar fields \cite{jin2020multi,ramesh2021multi}.

\section{Conclusions}
\label{sec:con}
This study introduces the novel research problem of temporal segmentation of full-procedure colonoscopy videos and provides the first extensive set of annotations for this task, along with a model training and evaluation protocol for this new benchmark. Furthermore, a state-of-the-art architecture utilizing a stack of custom temporal convolution blocks (ColonTCN) is presented and compared against competitive approaches. Ablation studies are also conducted to provide insights into the difficulties of this new task and the effectiveness of the selected ColonTCN model components. As automated colonoscopy analysis continues to evolve, we believe this work proposes an effective solution for this novel problem and can provide a robust foundation for future innovations in the field, aiming to enhance CAD(e/x) systems and ultimately improve colonoscopy screening effectiveness.

\section*{Data Availability}
The REAL-Colon dataset is released at \url{https://doi.org/10.25452/figshare.plus.22202866} while the frame-wise annotations released with this work have been released here \url{https://doi.org/10.6084/m9.figshare.26472913}. Code to automatically download and process the dataset has been made available at \url{https://github.com/cosmoimd/temporal_segmentation}.

\section*{Conflict of Interest Statement}
All authors are affiliated with Cosmo Intelligent Medical Devices, the developer of the GI Genius medical device.

\section*{Ethics statement}

This research study was conducted retrospectively using human subject data made available in open access by~\cite{biffi2024real}. Ethical approval was not required as confirmed by the license attached with the open-access data.

\section*{Acknowledgments}
We wish to express our gratitude to Antonella Melina and Michela Ruperti for their assistance in coordinating the data annotation process of the REAL-Colon dataset, and to the Data Annotation Team at Cosmo Intelligent Medical Devices for their diligent efforts in annotating the dataset.

\section*{Comparison without class weighting}
We report Table 2 of the paper obtained by training models without class weighting based on median frequency balancing.
The primary effect of not correctin for data imbalance is that models struggle to learn from less represented classes and tend to favor predictions for the majority classes, as expected, leading to an overall decrease in performance across all models. Specifically, the ileum class becomes almost impossible to learn. The model ranking remains consistent in both the 5-fold and 4-fold scenarios to what reported in Table 2.

\begin{table}[h!]
\caption{Comparison of the proposed ColonTCN approach with baseline architectures without class weighting at training.}
\centering
\begin{center}
\resizebox{\textwidth}{!}{
\begin{tabular}{@{}|ccc|ccccccccc|ccc|@{}}
\toprule
\multicolumn{15}{|c|}{5-fold REAL-Colon} \\
\midrule
Model & Params [M] & GFLOPs & Outside & Insertion & Ceacum & Ileum & Ascending & Transverse & Descending & Sigmoid & Rectum &  wF1 & wJacc & WMAPE \\
\midrule
TCN (TeCNO) & 0.5 & 2.809 & 86 & 62.1 & 37.2 & 12.7 & 33.8 & 34.7 & 21.2 & 35.1 & 37.5 & 44.1 & 29.7 & 20.1 \\
MS-TCN (TeCNO) & 1.7 & 9.351 & 82.5 & 63.6 & 37.4 & 20.5 & 33.6 & 33 & 23.4 & 35.5 & 37.4 & 44.6 & 30.1 & 19.7 \\
ASFormer & 1.2 & 6.376 & 14.7 & 84.6 & 24.4 & 0 & 31.8 & 54.4 & 19.2 & 38.1 & 29.8 & 53.5 & 40.9 & 23.8 \\
ColonTCN & 0.9 & 4.386 & 87.8 & 91.8 & 47.6 & 0 & 37.9 & 70.7 & 38.9 & 67.8 & 74 & 70.5 & 58.1 & 7.9 \\
\midrule
\multicolumn{15}{|c|}{4-fold REAL-Colon} \\
\midrule
Model & Params [M] & GFLOPs & Outside & Insertion & Ceacum & Ileum & Ascending & Transverse & Descending & Sigmoid & Rectum &  wF1 & wJacc & WMAPE \\
\midrule
TCN (TeCNO ) & 0.5 & 2.809 & 83 & 66.4 & 18.4 & 0 & 12.9 & 43.7 & 6.1 & 28.9 & 19.7 & 41.2 & 28.7 & 29.4 \\
MS-TCN (TeCNO ) & 1.7 & 9.351 & 24.7 & 75.5 & 9.9 & 0 & 25.7 & 33.3 & 16.7 & 28.1 & 28.2 & 43 & 30.9 & 21.6 \\
ASFormer & 1.2 & 6.376 & 5.2 & 85.8 & 29.7 & 0 & 29.1 & 42.9 & 27.9 & 27 & 31.1 & 51.2 & 39.1 & 16.7 \\
ColonTCN & 0.9 & 4.386 & 73.4 & 82.4 & 41.5 & 0 & 36.5 & 50.5 & 6.4 & 41 & 60.2 & 55.6 & 42.2 & 18.2 \\
\bottomrule
\end{tabular}}
\end{center}
\label{tbl:maintab}
\end{table}

\clearpage
\newpage
\clearpage

\section*{Video Distribution in 5-Fold Cross-Validation}

This section details the distribution of videos across the five folds of the cross-validation evaluation process adopted in the study. Each fold's division into training, validation, and testing sets is designed to ensure there are no overlaps between the sets, each video appears once in the test set, and a similar proportion of the ileum frames is maintained across all five testing folds.

\begin{table}[ht]
\centering
\caption{Detailed Video Distribution for Fold 1}
\label{tab:video-distribution-full}
\resizebox{\textwidth}{!}{%
\begin{tabular}{|c|p{6.5cm}|p{3cm}|p{6.5cm}|}
\hline
\textbf{Fold} & \textbf{Train Videos} & \textbf{Valid Videos} & \textbf{Test Videos} \\ \hline
1 & \begin{tabular}[t]{@{}l@{}}
\texttt{001-003, 001-004, 001-013} \\
\texttt{001-014, 001-005, 001-011} \\
\texttt{001-015, 001-006, 001-007} \\
\texttt{001-008, 001-012, 002-005} \\
\texttt{002-006, 002-007, 002-008} \\
\texttt{002-009, 002-010, 002-011} \\
\texttt{002-012, 002-013, 002-014} \\
\texttt{002-015, 003-005, 003-006} \\
\texttt{003-007, 003-008, 003-009} \\
\texttt{003-010, 003-011, 003-012} \\
\texttt{003-013, 003-014, 003-015} \\
\texttt{004-013, 004-004, 004-005} \\
\texttt{004-006, 004-007, 004-008} \\
\texttt{004-009, 004-010, 004-011} \\
\texttt{004-012, 004-015}
\end{tabular}
& \begin{tabular}[t]{@{}l@{}}
\texttt{001-010}, \\
\texttt{002-004}, \\
\texttt{003-004}, \\
\texttt{004-003} 
\end{tabular}
& \begin{tabular}[t]{@{}l@{}}
\texttt{001-009, 001-001, 001-002} \\
\texttt{002-001, 002-002, 002-003} \\
\texttt{003-001, 003-002, 003-003} \\
\texttt{004-001, 004-014, 004-002}
\end{tabular} \\ \hline
\end{tabular}%
}
\end{table}

\begin{table}[ht]
\centering
\caption{Detailed Video Distribution for Fold 2}
\label{tab:video-distribution-full}
\resizebox{\textwidth}{!}{%
\begin{tabular}{|c|p{6.5cm}|p{3cm}|p{6.5cm}|}
\hline
\textbf{Fold} & \textbf{Train Videos} & \textbf{Valid Videos} & \textbf{Test Videos} \\ \hline
2 & \begin{tabular}[t]{@{}l@{}}
\texttt{001-014, 001-005, 001-011} \\
\texttt{001-015, 001-006, 001-007} \\
\texttt{001-008, 001-012, 001-009} \\
\texttt{001-001, 001-002, 002-008} \\
\texttt{002-009, 002-010, 002-011} \\
\texttt{002-012, 002-013, 002-014} \\
\texttt{002-015, 002-001, 002-002} \\
\texttt{002-003, 003-008, 003-009} \\
\texttt{003-010, 003-011, 003-012} \\
\texttt{003-013, 003-014, 003-015} \\
\texttt{003-001, 003-002, 003-003} \\
\texttt{004-006, 004-007, 004-008} \\
\texttt{004-009, 004-010, 004-011} \\
\texttt{004-012, 004-015, 004-001} \\
\texttt{004-014, 004-002}
\end{tabular}
& \begin{tabular}[t]{@{}l@{}}
\texttt{001-013} \\
\texttt{002-007} \\
\texttt{003-007} \\
\texttt{004-005}
\end{tabular}
& \begin{tabular}[t]{@{}l@{}}
\texttt{001-010, 001-003, 001-004} \\
\texttt{002-004, 002-005, 002-006} \\
\texttt{003-004, 003-005, 003-006} \\
\texttt{004-003, 004-013, 004-004}
\end{tabular} \\ \hline
\end{tabular}
}
\end{table}

\begin{table}[ht]
\centering
\caption{Detailed Video Distribution for Fold 3}
\label{tab:video-distribution-fold3}
\resizebox{\textwidth}{!}{%
\begin{tabular}{|c|p{6.5cm}|p{3cm}|p{6.5cm}|}
\hline
\textbf{Fold} & \textbf{Train Videos} & \textbf{Valid Videos} & \textbf{Test Videos} \\ \hline
3 & \begin{tabular}[t]{@{}l@{}}
\texttt{001-015, 001-006, 001-007} \\
\texttt{001-008, 001-012, 001-009} \\
\texttt{001-001, 001-002, 001-010} \\
\texttt{001-003, 001-004, 002-011} \\
\texttt{002-012, 002-013, 002-014} \\
\texttt{002-015, 002-001, 002-002} \\
\texttt{002-003, 002-004, 002-005} \\
\texttt{002-006, 003-011, 003-012} \\
\texttt{003-013, 003-014, 003-015} \\
\texttt{003-001, 003-002, 003-003} \\
\texttt{003-004, 003-005, 003-006} \\
\texttt{004-009, 004-010, 004-011} \\
\texttt{004-012, 004-015, 004-001} \\
\texttt{004-014, 004-002, 004-003} \\
\texttt{004-013, 004-004}
\end{tabular}
& \begin{tabular}[t]{@{}l@{}}
\texttt{001-011} \\
\texttt{002-010} \\
\texttt{003-010} \\
\texttt{004-008}
\end{tabular}
& \begin{tabular}[t]{@{}l@{}}
\texttt{001-013, 001-014, 001-005} \\
\texttt{002-007, 002-008, 002-009} \\
\texttt{003-007, 003-008, 003-009} \\
\texttt{004-005, 004-006, 004-007}
\end{tabular} \\ \hline
\end{tabular}
}
\end{table}

\begin{table}[ht]
\centering
\caption{Detailed Video Distribution for Fold 4}
\label{tab:video-distribution-fold4}
\resizebox{\textwidth}{!}{%
\begin{tabular}{|c|p{6.5cm}|p{3cm}|p{6.5cm}|}
\hline
\textbf{Fold} & \textbf{Train Videos} & \textbf{Valid Videos} & \textbf{Test Videos} \\ \hline
4 & \begin{tabular}[t]{@{}l@{}}
\texttt{001-008, 001-012, 001-009} \\
\texttt{001-001, 001-002, 001-010} \\
\texttt{001-003, 001-004, 001-013} \\
\texttt{001-014, 001-005, 002-014} \\
\texttt{002-015, 002-001, 002-002} \\
\texttt{002-003, 002-004, 002-005} \\
\texttt{002-006, 002-007, 002-008} \\
\texttt{002-009, 003-014, 003-015} \\
\texttt{003-001, 003-002, 003-003} \\
\texttt{003-004, 003-005, 003-006} \\
\texttt{003-007, 003-008, 003-009} \\
\texttt{004-012, 004-015, 004-001} \\
\texttt{004-014, 004-002, 004-003} \\
\texttt{004-013, 004-004, 004-005} \\
\texttt{004-006, 004-007}
\end{tabular}
& \begin{tabular}[t]{@{}l@{}}
\texttt{001-007} \\
\texttt{002-013} \\
\texttt{003-013} \\
\texttt{004-011}
\end{tabular}
& \begin{tabular}[t]{@{}l@{}}
\texttt{001-011, 001-015, 001-006} \\
\texttt{002-010, 002-011, 002-012} \\
\texttt{003-010, 003-011, 003-012} \\
\texttt{004-008, 004-009, 004-010}
\end{tabular} \\ \hline
\end{tabular}
}
\end{table}

\begin{table}[ht]
\centering
\caption{Detailed Video Distribution for Fold 5}
\label{tab:video-distribution-fold5}
\resizebox{\textwidth}{!}{%
\begin{tabular}{|c|p{6.5cm}|p{3cm}|p{6.5cm}|}
\hline
\textbf{Fold} & \textbf{Train Videos} & \textbf{Valid Videos} & \textbf{Test Videos} \\ \hline
5 & \begin{tabular}[t]{@{}l@{}}
\texttt{001-001, 001-002, 001-010} \\
\texttt{001-003, 001-004, 001-013} \\
\texttt{001-014, 001-005, 001-011} \\
\texttt{001-015, 001-006, 002-002} \\
\texttt{002-003, 002-004, 002-005} \\
\texttt{002-006, 002-007, 002-008} \\
\texttt{002-009, 002-010, 002-011} \\
\texttt{002-012, 003-002, 003-003} \\
\texttt{003-004, 003-005, 003-006} \\
\texttt{003-007, 003-008, 003-009} \\
\texttt{003-010, 003-011, 003-012} \\
\texttt{004-014, 004-002, 004-003} \\
\texttt{004-013, 004-004, 004-005} \\
\texttt{004-006, 004-007, 004-008} \\
\texttt{004-009, 004-010}
\end{tabular}
& \begin{tabular}[t]{@{}l@{}}
\texttt{001-009} \\
\texttt{002-001} \\
\texttt{003-001} \\
\texttt{004-001}
\end{tabular}
& \begin{tabular}[t]{@{}l@{}}
\texttt{001-007, 001-008, 001-012} \\
\texttt{002-013, 002-014, 002-015} \\
\texttt{003-013, 003-014, 003-015} \\
\texttt{004-011, 004-012, 004-015}
\end{tabular} \\ \hline
\end{tabular}
}
\end{table}

\clearpage
\newpage
\clearpage

\section*{Exemplar REAL-Colon images}

\begin{figure}[ht!]
    \centering
    \includegraphics[width=0.7\textwidth]{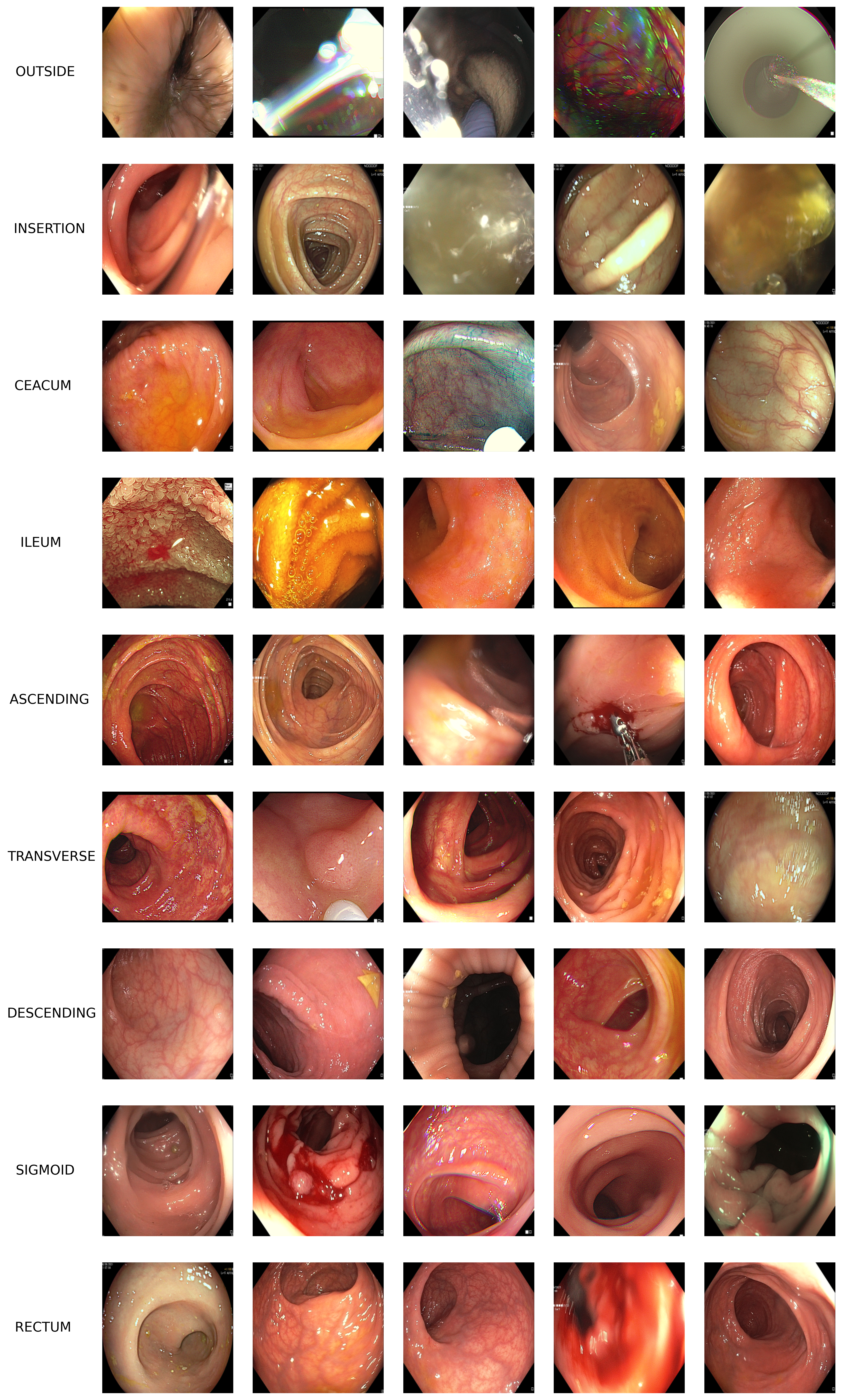}
    \caption{REAL-Colon dataset variability.}
    \label{fig:variability-visualization}
\end{figure}

In Figure \ref{fig:variability-visualization} we provide a high-resolution visualization of the REAL-Colon dataset images where each row corresponds to a unique class (GT value). Five random frames are displayed as columns for each class, showcasing the variability within the dataset. 

\clearpage
\newpage
\clearpage

\section*{Segments Duration}
Table \ref{tbl:duration} reports the absolute mean, minimum and maximum frame class annotation duration (in seconds) across all videos in the REAL-Colon dataset. 

\begin{table}[h!]
\footnotesize

\centering
\begin{center}
\begin{tabular}{|c|c|}
\hline
\textbf{Class Label} & \textbf{Mean [Min, Max] Duration [s]} \\ \hline
Outside & $29.1 \; [0, 105]$ \\ \hline
Insertion & $594.4 \; [113, 2608]$ \\ \hline
Ceacum & $133.2 \; [17, 608]$ \\ \hline
Ileum & $10.0 \; [0, 132]$\\ \hline
Ascending & $140.6 \; [13, 656]$ \\ \hline
Transverse & $355.0 \; [49, 1964]$ \\ \hline
Descending & $148.2 \; [13, 1132] $\\ \hline
Sigmoid & $176.8 \; [25, 726] $\\ \hline
Rectum & $101.4 \; [5, 597] $\\ \hline
Uncertain & $2.3 \; [0, 131] $ \\ \hline
\hline
Withdrawal & $1065.2 \; [294, 3256] $ \\ \hline
Procedure & $2753.8 \; [1092, 7483] $ \\ \hline
\end{tabular}
\caption{Mean, maximum and minimum frame class annotation duration (in seconds) across the 60 REAL-Colon videos. The final two rows, "Withdrawal" and "Procedure," aggregate durations for ceacum, ileum, ascending, transverse, descending, sigmoid, and rectum segments during withdrawal, and for withdrawal and insertion frames during the procedure, respectively. Note that "Procedure" represents the total time spent within the colon.}
\label{tbl:duration}
\end{center}
\end{table}

\bibliography{mybibfile}

\end{document}

%% file: fig_text/fig1a.tex
\begin{figure*}[t]
\centering
\caption{Center: Distribution of frame labels in the REAL-Colon dataset. On the left: Graphical visualization of the outside and insertion phases. On the right: Graphical visualization of the outside and withdrawal phases, further segmented into seven anatomical locations.}
\includegraphics[width=0.85\linewidth]{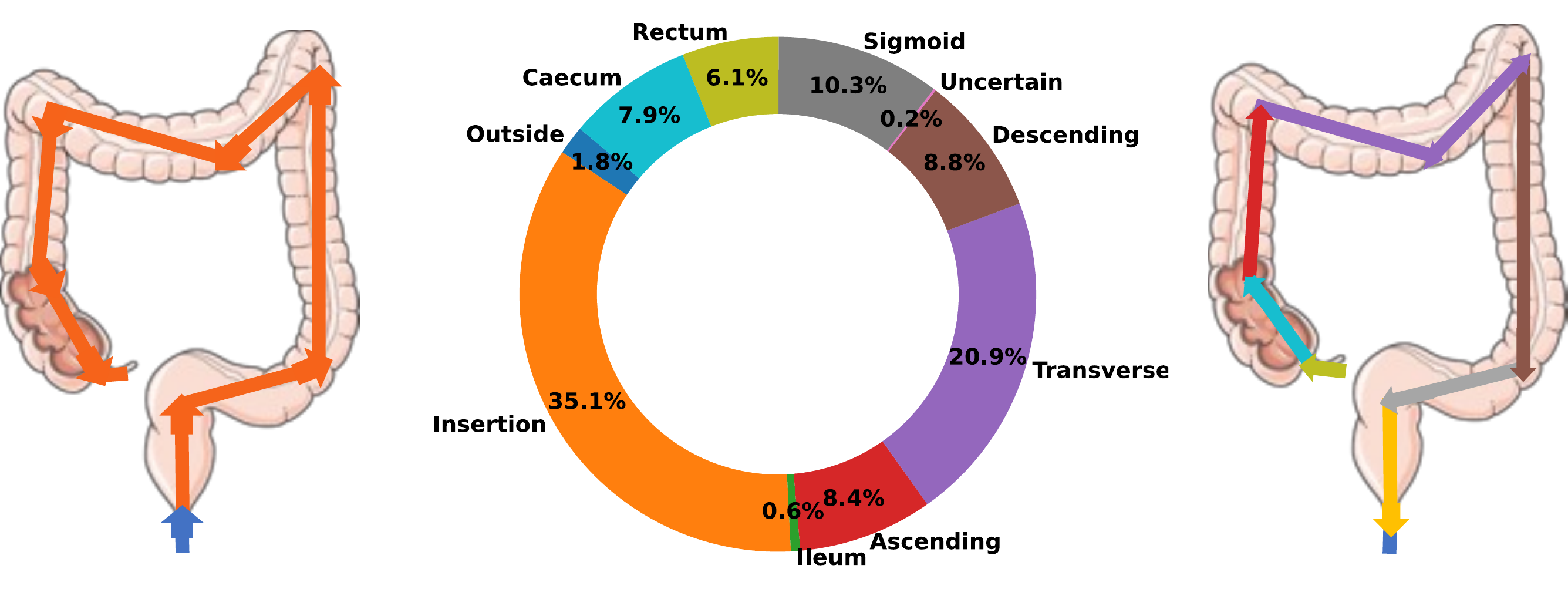}
\label{fig:fig1a}
\end{figure*}

%% file: fig_text/fig1b.tex
\begin{figure*}[t]
\centering
\caption{Box and whisker plot of the percentage of video length occupied by each frame label in the REAL-Colon dataset.}
\includegraphics[width=0.85\linewidth]{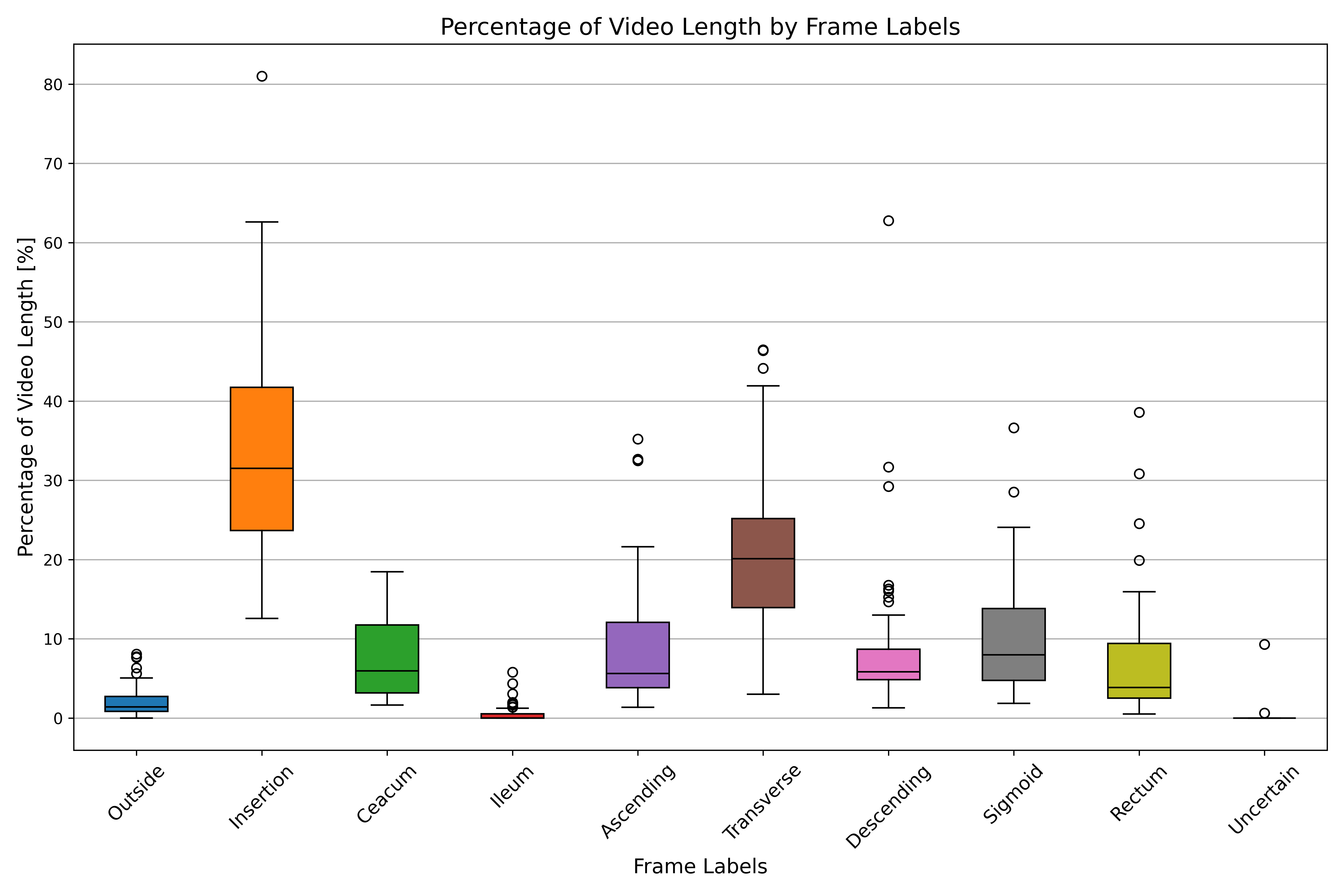}
\label{fig:fig1b}
\end{figure*}

%% file: fig_text/fig2a.tex
\begin{figure*}[t!]
\centering
\caption{Overview of the proposed ColonTCN approach for colonoscopy video temporal segmentation.}
\includegraphics[width=\linewidth]{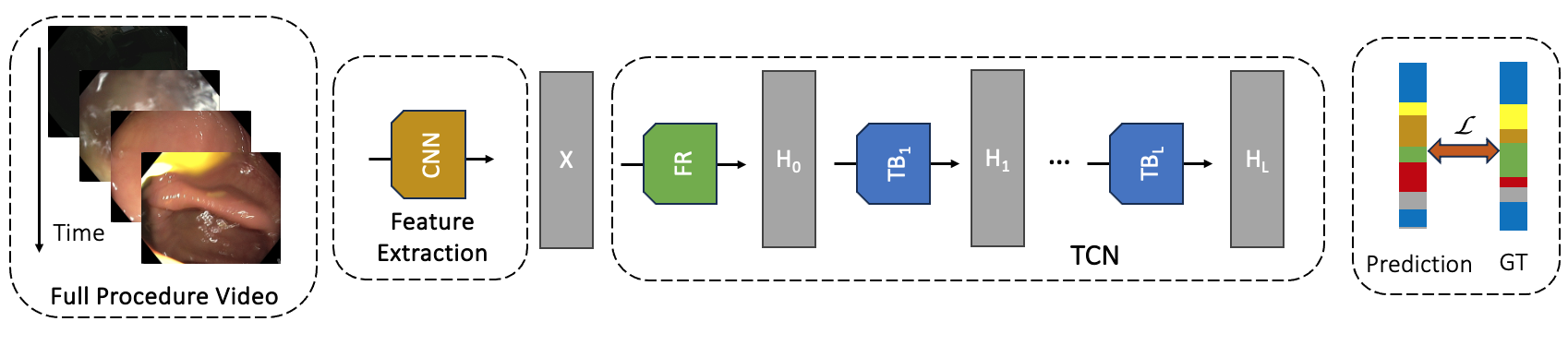}
\label{fig:fig2a}
\end{figure*}

%% file: fig_text/fig2b.tex
\begin{figure}[t!]
\centering
\caption{Overview of the Temporal Block (TB) used in the proposed ColonTCN approach.}
\includegraphics[width=0.6\linewidth]{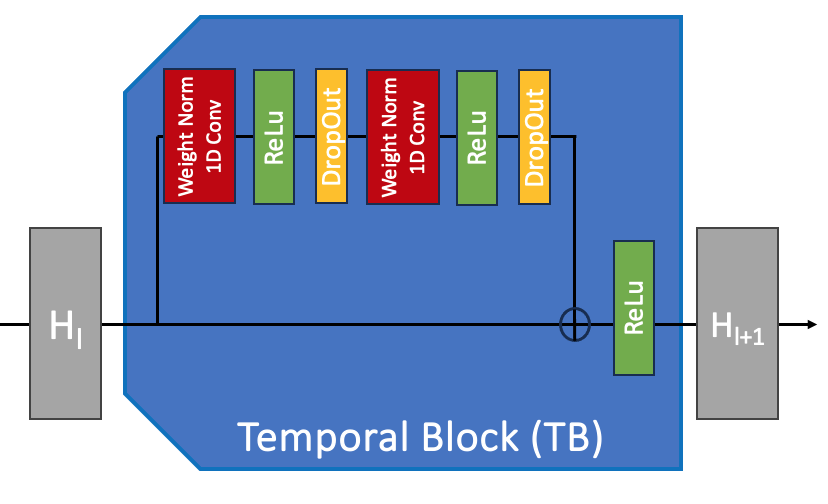}
\label{fig:fig2b}
\end{figure}

%% file: tables/folds.tex
\begin{table}[!htbp]
\caption{Test set class distribution for 5-fold and 4-fold evaluations on the REAL-Colon dataset.}
\centering
\begin{center}
\resizebox{\textwidth}{!}{
\begin{tabular}{@{}|cc|ccccccccc|c|@{}}
\toprule
\multicolumn{12}{|c|}{5-fold REAL-Colon} \\
\midrule
Fold & Cohort & Outside & Insertion & Ceacum & Ileum & Ascending & Transverse & Descending & Sigmoid & Rectum &  Total \\
\midrule
 1 & 1,2,3,4 & 1,600 & 26,622 & 5,357 & 503 & 12,321 & 25,548 & 7,236 & 14,046 & 3,082 & 96,315\\
 2 & 1,2,3,4 & 1,886 & 51,360 & 9,638 & 783 & 7,724 & 26,079 & 11,636 & 10,707 & 6,168 & 125,981  \\
 3 & 1,2,3,4 & 1,553 & 23,779 & 6,632 & 261 & 8,359 & 19,033 & 7,974 & 10,540 & 4,929 & 83,060 \\
 4 & 1,2,3,4 & 1,472 & 30,881 & 9,393 & 944 & 7,849 & 15,332 & 5,666 & 9,862 & 9,625 & 91,024 \\
 5 & 1,2,3,4 & 2,372 & 45,761 & 9,060 & 557 & 6,048 & 20,613 & 12,106 & 8,007 & 6,740 & 111,264\\
\midrule
Total & & 8,883 & 178,403 & 40,080 & 3,048 & 42,301 & 106,605 & 44,618 & 53,162 & 30,544 & 507,644  \\
\midrule
\midrule
\multicolumn{12}{|c|}{4-fold REAL-Colon} \\
\midrule
Fold & Cohort & Outside & Insertion & Ceacum & Ileum & Ascending & Transverse & Descending & Sigmoid & Rectum &  Total \\
\midrule
 1 & 4 & 1,310 & 31,933 & 3,942 & 436 & 6,915 & 18,424 & 10,250 & 5,091 & 7,605 & 85,906 \\
 2 & 3 & 2,158 & 77,646 & 21,088 & 2,054 & 16,882 & 53,805 & 21,874 & 26,825 & 9,041 & 231,373  \\
 3 & 2 & 1,881 & 38,840 & 4,775 & 0 & 7,409 & 15,369 & 7,013 & 14,216 & 10,293 & 99,796 \\
 4 & 1 & 3,534 & 29,984 & 10,275 & 558 & 11,095 & 19,007 & 5,481 & 7,030 & 3,605 & 90,569 \\
\midrule
Total & & 8,883 & 178,403 & 40,080 & 3,048 & 42,301 & 106,605 & 44,618 & 53,162 & 30,544 & 507,644  \\
\bottomrule
\end{tabular}}
\end{center}
\label{tbl:folds}
\end{table}

%% file: tables/main_tab.tex
\begin{table}[t]
\caption{Comparison of the proposed ColonTCN approach with baseline architectures. For each model, the F1 score for each class, the wF1, the wJacc and the WMAPE computed on all the 60 REAL-Colon dataset videos, and the number of parameters and GFLOPs is reported.}
\centering
\begin{center}
\resizebox{\textwidth}{!}{
\begin{tabular}{@{}|ccc|ccccccccc|ccc|@{}}
\toprule
\multicolumn{15}{|c|}{5-fold REAL-Colon} \\
\midrule
Model & Params [M] & GFLOPs & Outside & Insertion & Ceacum & Ileum & Ascending & Transverse & Descending & Sigmoid & Rectum &  wF1 & wJacc & WMAPE \\
\midrule
TCN (TeCNO \cite{czempiel2020tecno}) & 0.5 & 2.809 & \(81.9\) & \(85.8\) & \(40.3\) & \(4.1\) & \(34.8\) & \(58.5\) & \(35.1\) & \(58.4\) & \(63.4\) & \(63.0\) & \(49.0\) & \(13.5\)  \\
MS-TCN (TeCNO \cite{czempiel2020tecno}) & 1.7 & 9.351 & \(87.9\) & \(81.9\) & \(43.7\) & \(27.4\) & \(33.3\) & \(5.04\) & \(19.7\) & \(45.7\) & \(66.0\)  & \(58.5\)& \(44.5\)  & \(21.4\) \\
ASFormer \cite{yi2021asformer} & 1.2 & 6.376 & \(\textbf{96.2}\) & \(95.1\) & \(63.8\) & \(52.1\) & \(\textbf{59.1}\) & \(68.9\) & \(\textbf{42.8}\) & \(\textbf{67.0}\) & \(\textbf{77.4} \) & \(75.2\) & \(63.4\) & \(\textbf{3.3}\)  \\
ColonTCN (Ours) & 0.9 & 4.386 & \(95.0\) & \(\textbf{95.9}\) & \(\textbf{68.0}\) & \(\textbf{56.2}\) & \(49.8\) & \(\textbf{75.2}\) & \(47.6\) & \(65.6\) & \(72.1\) & \(\textbf{76.4}\) & \(\textbf{65.0}\) & \(4.3\) \\

\midrule
\midrule
\multicolumn{15}{|c|}{4-fold REAL-Colon} \\
\midrule
Model & Params [M] & GFLOPs & Outside & Insertion & Ceacum & Ileum & Ascending & Transverse & Descending & Sigmoid & Rectum &  wF1 & wJacc & WMAPE \\
\midrule
TCN (TeCNO \cite{czempiel2020tecno}) & 0.5 & 2.809 &  \(73.6\) & \(71.6\) & \(32.0\) & \(0.0\) & \(24.9\) & \(35.8\) & \(22.6\) & \(34.4\) & \(60.5\) & \(47.8\) & \(33.8\) & \(24.8\) \\
MS-TCN (TeCNO \cite{czempiel2020tecno}) & 1.7 & 9.351 & \(87.5\) & \(83.4\) & \(39.5\) & \(10.9\) & \(32.9\) & \(44.5\) & \(18.1\) & \(49.5\) & \(60.8\) & \(56.5\) & \(43.1\) & \(18.6\) \\
ASFormer \cite{yi2021asformer} & 1.2 & 6.376 & \(83.9\) & \(\textbf{86.7}\) & \(45.9\) & \(30.8\) & \(41.7\) & \(54.2\) & \(27.6\) & \(40.0\) & \(63.4\) & \(61.0\) & \(47.4\) & \(16.8\)  \\
ColonTCN (Ours) & 0.8 & 4.099 & \(\textbf{88.1}\) & \(85.2\) & \(\textbf{51.2}\) & \(\textbf{31.1}\) & \(\textbf{41.6}\) & \(\textbf{56.6}\) & \(\textbf{25.6}\) & \(\textbf{53.5}\) & \(\textbf{66.6}\) & \(\textbf{62.9}\) & \(\textbf{48.9}\) & \(\textbf{16.2}\) \\
\bottomrule
\end{tabular}}
\end{center}
\label{tbl:maintab}
\end{table}

%% file: tables/besttcn.tex
\begin{table}[!htbp]
\caption{Ablation study: Impact of feature reduction (FR) and residual dilated convolutions, double convolutions and dropout in each TB on the best performing ColonTCN model.}
\centering
\begin{center}
\resizebox{\textwidth}{!}{
\begin{tabular}{@{}|cccc|ccccccccc|ccc|@{}}
\toprule
\multicolumn{16}{|c|}{5-fold REAL-Colon} \\
\midrule
FR & DropOut & Double & Residual & Outside & Insertion & Ceacum & Ileum & Ascending & Transverse & Descending & Sigmoid & Rectum &  wF1 & wJacc & WMAPE \\
\midrule
\checkmark &  \checkmark & \checkmark & \checkmark & \(95.0\) & \(\textbf{95.9}\) & \(\textbf{68.0}\) & \(\textbf{56.2}\) & \(\textbf{49.8}\) & \(\textbf{75.2}\) & \(\textbf{47.6}\) & \(\textbf{65.6}\) & \(72.1\) & \(\textbf{76.4}\) & \(\textbf{65.0}\) & \(\textbf{4.3}\) \\
$\times$ & \checkmark & \checkmark & \checkmark & \(\textbf{96.2}\) & \(92.8\) & \(57.3\) & \(30.1\) & \(41.9\) & \(65.5\) & \(40.5\) & \(\textbf{65.6}\) & \(\textbf{72.8}\) & \(71\) & \(58.5\)  & \(9.2\)\\
\checkmark & $\times$ & \checkmark & \checkmark & \(\textbf{96.2}\) & \(93.8\) & \(65.1\) & \(50.3\) & \(45.0\) & \(63.7\) & \(39.0\) & \(59.0\) & \(68.3\) & \(70.9\) & \(58.5\) & \(7.5\) \\
\checkmark & \checkmark & $\times$ & \checkmark & \(95.2\) & \(91.3\) & \(52.3\) & \(19.3\) & \(43.9\) & \(68.4\) & \(34.8\) & \(59.8\) & \(72.8\) & \(69.7\)  & \(57.0\) & \(14.5\) \\
\checkmark & \checkmark & \checkmark & $\times$ & \(95.8\) & \(92.1\) & \(57.2\) & \(8.9\) & \(45.4\) & \(67.3\) & \(33.0\) & \(59.0\) & \(73.1\) & \(70.0\) & \(57.5\) & \(12.3\)  \\ 
\checkmark & \checkmark & $\times$ & $\times$ & \(81.9\) & \(85.8\) & \(40.3\) & \(4.1\) & \(34.8\) & \(58.5\) & \(35.1\) & \(58.4\) & \(63.4\) & \(63.0\) & \(49.0\) & \(13.5\)  \\
\midrule
\midrule
\multicolumn{16}{|c|}{4-fold REAL-Colon} \\
\midrule
FR & DropOut & Double & Residual & Outside & Insertion & Ceacum & Ileum & Ascending & Transverse & Descending & Sigmoid & Rectum &  wF1 & wJacc & WMAPE \\
\midrule
\checkmark &  \checkmark & \checkmark & \checkmark &\(88.1\) & \(85.2\) & \(\textbf{51.2}\) & \(\textbf{31.1}\) & \(\textbf{41.6}\) & \(\textbf{56.6}\) & \(25.6\) & \(\textbf{53.5}\) & \(\textbf{66.6}\) & \(\textbf{62.9}\) & \(\textbf{48.9}\) & \(16.2\)  \\
$\times$ & \checkmark & \checkmark & \checkmark & \(\textbf{91.1}\) & \(\textbf{86.3}\) & \(3.7\) & \(3.1\) & \(28.2\) & \(51.7\) & \(12.9\) & \(38.8\) & \(65.1\) & \(57.2\) & \(44.7\) & \(\textbf{12.8}\) \\
\checkmark & $\times$ & \checkmark & \checkmark & \(78.7\) & \(83.5\) & \(32.2\) & \(12.0\) & \(38.5\) & \(47.6\) & \(17.1\) & \(39.4\) & \(52.5\) & \(55.3\) & \(41.9\) & \(18.0\) \\
\checkmark & \checkmark & $\times$ & \checkmark & \(73.2\) & \(78.8\) & \(21.2\) & \(1.9\) & \(26.3\) & \(37.9\) & \(\textbf{30.4}\) & \(34.4\) & \(52.6\) & \(50.3\) & \(36.9\) & \(24.7\) \\
\checkmark & \checkmark & \checkmark & $\times$ & \(87.4\) & \(81.8\) & \(35.7\) & \(10.7\) & \(36.4\) & \(47.0\) & \(15.8\) & \(50.2\) & \(56.1\) & \(56.1\) & \(42.4\) & \(18.9\) \\
\checkmark & \checkmark & $\times$ & $\times$ & \(73.6\) & \(71.6\) & \(32.0\) & \(0.0\) & \(24.9\) & \(35.8\) & \(22.6\) & \(34.4\) & \(60.5\) & \(47.8\) & \(33.8\) & \(24.8\) \\
\bottomrule
\end{tabular}}
\end{center}
\label{tbl:besttcn}
\end{table}

%% file: tables/levels.tex
\begin{table}[t]
\caption{ColonTCN model performance as a function of the TB count (levels).}
\centering
\begin{center}
\resizebox{\textwidth}{!}{
\begin{tabular}{@{}|c|ccccccccc|ccc|@{}}
\toprule
\multicolumn{13}{|c|}{5-fold REAL-Colon} \\
\midrule
Levels & Outside & Insertion & Ceacum & Ileum & Ascending & Transverse & Descending & Sigmoid & Rectum &  wF1 & wJacc & WMAPE \\
\midrule
10 & \(\textbf{96.2}\) & \(93.8\) & \(65.1\) & \(50.3\) & \(45.0\) & \(63.7\) & \(39.0\) & \(59.0\) & \(68.3\) & \(70.9\) & \(58.5\) & \(7.5\) \\
11 & \(95.6\) & \(94.1\) & \(60.7\) & \(47.0\) & \(46.1\) & \(65.2\) & \(41.3\) & \(60.4\) & \(68.3\) & \(71.4\)  & \(59.0\) & \(6.6\) \\
12 & \(\textbf{96.2}\) & \(95.7\) & \(66.3\) & \(55.0\) & \(51.7\) & \(68.6\) & \(43.8\) & \(64.8\) & \(71.3\) & \(74.5\)  & \(62.7\) & \(5.4\) \\
13 &  \(95.0\) & \(\textbf{95.9}\) & \(\textbf{68.0}\) & \(\textbf{56.2}\) & \(49.8\) & \(\textbf{75.2}\) & \(47.6\) & \(65.6\) & \(72.1\) & \(\textbf{76.4}\) & \(\textbf{65.0}\) & \(\textbf{4.3}\) \\
14 & \(95.1\) & \(94.4\) & \(64.5\) & \(52.4\) & \(\textbf{51.8}\) & \(73.2\) & \(\textbf{48.4}\) & \(\textbf{68.1}\) & \(\textbf{74.2}\) & \(75.8\)  & \(63.8\) & \(7.1\)\\
\midrule
\midrule
\multicolumn{13}{|c|}{4-fold REAL-Colon} \\
\midrule
Levels & Outside & Insertion & Ceacum & Ileum & Ascending & Transverse & Descending & Sigmoid & Rectum &  wF1 & wJacc & WMAPE \\
\midrule
11 & \(87.9\) & \(81.9\) & \(43.7\) & \(27.4\) & \(33.3\) & \(54.0\) & \(19.7\) & \(45.7\) & \(66.0\)  & \(58.5\)& \(44.5\)  & \(21.4\) \\
12 &  \(88.1\) & \(\textbf{85.2}\) & \(\textbf{51.2}\) & \(\textbf{31.1}\) & \(\textbf{41.6}\) & \(\textbf{56.6}\) & \(\textbf{25.6}\) & \(\textbf{53.5}\) & \(66.6\) & \(\textbf{62.9}\) & \(\textbf{48.9}\) & \(\textbf{16.2}\)  \\
13 & \(\textbf{88.2}\) & \(83.4\) & \(47.5\) & \(28.7\) & \(38.5\) & \(51.9\) & \(15.5\) & \(47.7\) & \(\textbf{67.1}\) & \(59.3\) & \(45.5\) & \(20.1\)  \\
\bottomrule
\end{tabular}}
\end{center}
\label{tbl:levels}
\end{table}

%% file: tables/stages.tex
\begin{table}[t]
\caption{MS-ColonTCN performance assessment. The Levels column uses the first number to denote the number of TBs in the base TCN model, whilst the second number (when present) to indicate the number of TBs in the refinement stages. The Stages column instead indicates the number of refinement stages used.}
\centering
\begin{center}
\resizebox{\textwidth}{!}{
\begin{tabular}{@{}|cc|ccccccccc|ccc|@{}}
\toprule
\multicolumn{14}{|c|}{5-fold REAL-Colon} \\
\midrule
Levels & Stages & Outside & Insertion & Ceacum & Ileum & Ascending & Transverse & Descending & Sigmoid & Rectum &  wF1 & wJacc & WMAPE \\
\midrule
13 & 0 & \(95.0\) & \(95.9\) & \(68.0\) & \(56.2\) & \(49.8\) & \(75.2\) & \(47.6\) & \(65.6\) & \(72.1\) & \(6.4\) & \(65.0\) & \(4.3\) \\
\midrule
13-10 & 3 & \(96.5\) & \(93.5\) & \(61.8\) & \(36.1\) & \(45.8\) & \(69.4\) & \(42.8\) & \(65.7\) & \(73.0\) & \(73.1\) & \(60.0\) & \(6.5\) \\
13-11 & 3 & \(96.7\) & \(\textbf{94.2}\) & \(62.4\) & \(25.9\) & \(46.8\) & \(\textbf{74.4}\) & \(\textbf{47.7}\) & \(\textbf{69.4}\) & \(73.8\) & \(\textbf{75.3}\) & \(\textbf{63.4}\) & \(\textbf{5.0}\) \\
13-12 & 3 & \(93.1\) & \(93.3\) & \(60.9\) & \(\textbf{48.7}\) & \(\textbf{49.6}\) & \(72.1\) & \(43.0\) & \(66.9\) & \(74.5\) & \(74.1\) & \(61.7\) & \(6.8\) \\
13-13 & 3 & \(\textbf{96.9}\) & \(94.0\) & \(\textbf{62.7}\) & \(32.9\) & \(46.5\) & \(71.5\) & \(45.4\) & \(65.5\) & \(\textbf{75.4}\) & \(74.1\) & \(62.0\) & \(7.6\) \\
\midrule
13-11 & 1 & \(96.6\) & \(93.8\) & \(63.3\) & \(\textbf{47.0}\) & \(\textbf{46.8}\) & \(71.9\) & \(44.6\) & \(66.6\) & \(73.9\) & \(74.3\) & \(62.1\) & \(7.6\) \\
13-11 & 2 & \(96.0\) & \(94.0\) & \(\textbf{64.1}\) & \(42.4\) & \(46.3\) & \(71.6\) & \(45.7\) & \(65.9\) & \(\textbf{74.0}\) & \(74.3\) & \(62.2\) & \(7.2\) \\
13-11 & 3 & \(96.7\) & \(\textbf{94.2}\) & \(62.4\) & \(25.9\) & \(\textbf{46.8}\) & \(\textbf{74.4}\) & \(\textbf{47.7}\) & \(\textbf{69.4}\) & \(73.8\) & \(\textbf{75.3}\) & \(\textbf{63.4}\) & \(\textbf{5.0}\) \\
13-11 & 4 & \(\textbf{97.1}\) & \(92.5\) & \(56.0\) & \(46.5\) & \(45.3\) & \(71.4\) & \(42.9\) & \(64.4\) & \(69.3\) & \(72.4\) & \(59.8\) & \(6.5\) \\
\midrule
\midrule
\multicolumn{14}{|c|}{4-fold REAL-Colon} \\
\midrule
Levels & Stages & Outside & Insertion & Ceacum & Ileum & Ascending & Transverse & Descending & Sigmoid & Rectum &  wF1 & wJacc & WMAPE \\
\midrule
12 & 0 & \(88.1\) & \(85.2\) & \(51.2\) & \(31.1\) & \(41.6\) & \(56.6\) & \(25.6\) & \(53.5\) & \(66.6\) & \(62.9\) & \(48.9\) & \(16.2\)  \\
\midrule
12-10 & 3 & \(92.2\) & \(81.8\) & \(\textbf{40.7}\) & \(\textbf{19.3}\) & \(\textbf{46.3}\) & \(54.0\) & \(15.0\) & \(\textbf{50.2}\) & \(\textbf{72.2}\) & \(59.8\) & \(45.8\) & \(20.0\) \\
12-11 & 3 & \(\textbf{93.3}\) & \(\textbf{83.9}\) & \(38.1\) & \(9.8\) & \(41.3\) & \(\textbf{56.5}\) & \(\textbf{20.5}\) & \(49.3\) & \(62.2\) & \(\textbf{60.2}\) & \(\textbf{46.4}\) & \(16.1\) \\
12-12 & 3 & \(89.5\) & \(85.8\) & \(36.6\) & \(12.2\) & \(37.4\) & \(47.7\) & \(16.6\) & \(46.1\) & \(64.4\) & \(58.0\) & \(44.9\) & \(\textbf{15.0}\) \\
\midrule
12-11 & 1 & \(81.7\) & \(82.5\) & \(38.4\) & \(8.4\) & \(32.8\) & \(46.2\) & \(\textbf{21.2}\) & \(47.5\) & \(65.6\) & \(56.7\) & \(43.0\) & \(21.6\) \\
12-11 & 2 & \(91.0\) & \(\textbf{86.4}\) & \(\textbf{42.1}\) & \(\textbf{17.5}\) & \(34.5\) & \(47.4\) & \(16.6\) & \(46.0\) & \(\textbf{69.5}\) & \(58.7\) & \(45.7\) & \(\textbf{14.3}\) \\
12-11 & 3 & \(\textbf{93.3}\) & \(83.9\) & \(38.1\) & \(9.8\) & \(\textbf{41.3}\) & \(\textbf{56.5}\) & \(20.5\) & \(49.3\) & \(62.1\) & \(\textbf{60.2}\) & \(\textbf{46.4}\) & \(16.1\) \\
12-11 & 4 & \(87.5\) & \(83.4\) & \(39.5\) & \(10.9\) & \(32.9\) & \(44.5\) & \(18.1\) & \(\textbf{49.5}\) & \(60.8\) & \(56.5\) & \(43.1\) & \(18.6\) \\
\midrule
\bottomrule
\end{tabular}}
\end{center}
\label{tbl:stages}
\end{table}

%% file: tables/abl_aug.tex
\begin{table}[t]
\footnotesize
\centering
\caption{Effect of spatial and temporal data augmentation on ColonTCN, 5-fold CV setting.}
\begin{center}
\centering
\begin{tabular}{cccccc} 
\toprule
Spatial & Temporal & Ileum &  wF1 & wJacc & WMAPE  \\ 
\cmidrule(lr){1-1}\cmidrule(lr){2-2}\cmidrule(lr){3-3}\cmidrule(lr){4-4}\cmidrule(lr){5-5}\cmidrule(lr){6-6}
$\times$ & $\times$ & 44.1 & 73.9 & 61.4 & 5.3  \\
\checkmark   & $\times$  & 42.1 & 74.3  & 62.2 & 6.2  \\
$\times$   & \checkmark  & 51.2  & 74.6  & 62.8 & 5.2 \\
\checkmark  & \checkmark  & \textbf{56.2} & \textbf{76.4}  & \textbf{65.0} & \textbf{4.3} \\
\bottomrule
\end{tabular}
\label{tbl:abl_aug}
\end{center}
\end{table}

%% file: tables/losses.tex
\begin{table}[t]
\footnotesize

\caption{Comparison of loss functions contributions to the best ColonTCN model, 5-fold CV setting.}
\centering
\begin{center}
\centering
\begin{tabular}{ccccccc} 
\toprule
wCE & TMSE & Focal & Ileum &  wF1 & wJacc & WMAPE  \\ 
\cmidrule(lr){1-1}\cmidrule(lr){2-2}\cmidrule(lr){3-3}\cmidrule(lr){4-4}\cmidrule(lr){5-5}\cmidrule(lr){6-6}\cmidrule(lr){7-7}
\checkmark  &  $\times$  & $\times$  & 51.1 & 74.4 & 62.4 & 5.2  \\
\checkmark  & \checkmark   & $\times$  & \textbf{56.2} & \textbf{76.4}  & \textbf{65.0} & \textbf{4.3} \\
\checkmark  & $\times$   & \checkmark  & 51.2  & 73.1  & 60.8 & 6.5 \\
\checkmark  & \checkmark  & \checkmark  & 46.3  & 74.1  & 62.7 & 5.3  \\
\bottomrule
\end{tabular}
\end{center}
\label{tbl:losses}
\end{table}

%% file: fig_text/fig3.tex
\begin{figure*}[t]
\centering
\caption{Visualization of temporal segmentation on 6 random videos of the REAL-Colon dataset for ColonTCN and the best MS-ColonTCN models versus ASFormer prediction. GT indicates ground truth annotations. Models were trained in the 5-fold CV scenario.}
\includegraphics[width=\linewidth]{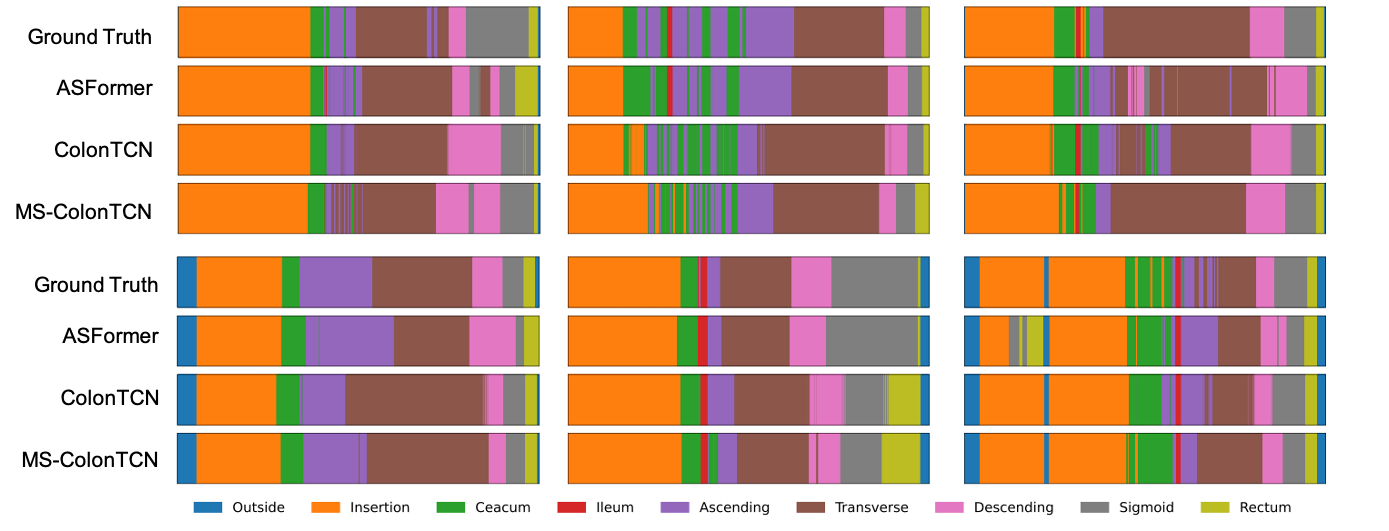}
\label{fig:fig3}
\end{figure*}

%% file: main.bbl
\begin{thebibliography}{10}

\bibitem{berzin_position_2020}
Tyler~M. Berzin, Sravanthi Parasa, Michael~B. Wallace, Seth~A. Gross,
  Alessandro Repici, and Prateek Sharma.
\newblock Position statement on priorities for artificial intelligence in {GI}
  endoscopy: a report by the {ASGE} {Task} {Force}.
\newblock {\em Gastrointestinal Endoscopy}, 92(4):951--959, October 2020.

\bibitem{biffi2024real}
Carlo Biffi, Giulio Antonelli, Sebastian Bernhofer, Cesare Hassan, Daizen
  Hirata, Mineo Iwatate, Andreas Maieron, Pietro Salvagnini, and Andrea
  Cherubini.
\newblock Real-colon: A dataset for developing real-world ai applications in
  colonoscopy.
\newblock {\em Scientific Data}, 11(1):539, 2024.

\bibitem{biffi_novel_2022}
Carlo Biffi, Pietro Salvagnini, Nhan Ngo~Dinh, Cesare Hassan, Prateek Sharma,
  {GI Genius CADx Study Group}, and Andrea Cherubini.
\newblock A novel {AI} device for real-time optical characterization of
  colorectal polyps.
\newblock {\em NPJ digital medicine}, 5(1):84, June 2022.

\bibitem{chang2022development}
Yuan-Yen Chang, Pai-Chi Li, Ruey-Feng Chang, Yu-Yao Chang, Siou-Ping Huang,
  Yang-Yuan Chen, Wen-Yen Chang, and Hsu-Heng Yen.
\newblock Development and validation of a deep learning-based algorithm for
  colonoscopy quality assessment.
\newblock {\em Surgical Endoscopy}, 36(9):6446--6455, 2022.

\bibitem{czempiel2020tecno}
Tobias Czempiel, Magdalini Paschali, Matthias Keicher, Walter Simson, Hubertus
  Feussner, Seong~Tae Kim, and Nassir Navab.
\newblock Tecno: Surgical phase recognition with multi-stage temporal
  convolutional networks.
\newblock In {\em Medical Image Computing and Computer Assisted
  Intervention--MICCAI 2020: 23rd International Conference, Lima, Peru, October
  4--8, 2020, Proceedings, Part III 23}, pages 343--352. Springer, 2020.

\bibitem{czempiel2021opera}
Tobias Czempiel, Magdalini Paschali, Daniel Ostler, Seong~Tae Kim, Benjamin
  Busam, and Nassir Navab.
\newblock Opera: Attention-regularized transformers for surgical phase
  recognition.
\newblock In {\em Medical Image Computing and Computer Assisted
  Intervention--MICCAI 2021: 24th International Conference, Strasbourg, France,
  September 27--October 1, 2021, Proceedings, Part IV 24}, pages 604--614.
  Springer, 2021.

\bibitem{de2023automated}
Thomas De~Carvalho, Rawen Kader, Patrick Brandao, Juana Gonz{\'a}lez-Bueno
  Puyal, Laurence~B Lovat, Peter Mountney, and Danail Stoyanov.
\newblock Automated colonoscopy withdrawal phase duration estimation using
  cecum detection and surgical tasks classification.
\newblock {\em Biomedical Optics Express}, 14(6):2629--2644, 2023.

\bibitem{dekker_advances_2018}
Evelien Dekker and Douglas~K. Rex.
\newblock Advances in {CRC} {Prevention}: {Screening} and {Surveillance}.
\newblock {\em Gastroenterology}, 154(7):1970--1984, May 2018.

\bibitem{demir2023deep}
Kubilay~Can Demir, Hannah Schieber, TobiasDaniel WeiseRoth, Matthias May,
  Andreas Maier, and Seung~Hee Yang.
\newblock Deep learning in surgical workflow analysis: A review of phase and
  step recognition.
\newblock {\em IEEE Journal of Biomedical and Health Informatics}, 2023.

\bibitem{ding2023temporal}
Guodong Ding, Fadime Sener, and Angela Yao.
\newblock Temporal action segmentation: An analysis of modern techniques.
\newblock {\em IEEE Transactions on Pattern Analysis and Machine Intelligence},
  2023.

\bibitem{ding2022exploring}
Xinpeng Ding and Xiaomeng Li.
\newblock Exploring segment-level semantics for online phase recognition from
  surgical videos.
\newblock {\em IEEE Transactions on Medical Imaging}, 41(11):3309--3319, 2022.

\bibitem{farha2019ms}
Yazan~Abu Farha and Jurgen Gall.
\newblock Ms-tcn: Multi-stage temporal convolutional network for action
  segmentation.
\newblock In {\em Proceedings of the IEEE/CVF conference on computer vision and
  pattern recognition}, pages 3575--3584, 2019.

\bibitem{feng2023development}
Lina Feng, Jiaxin Xu, Xuantao Ji, Liping Chen, Shuai Xing, Bo~Liu, Jian Han,
  Kai Zhao, Junqi Li, Suhong Xia, et~al.
\newblock Development and validation of a three-dimensional deep learning-based
  system for assessing bowel preparation on colonoscopy video.
\newblock {\em Frontiers in Medicine}, 10, 2023.

\bibitem{gao2021trans}
Xiaojie Gao, Yueming Jin, Yonghao Long, Qi~Dou, and Pheng-Ann Heng.
\newblock Trans-svnet: Accurate phase recognition from surgical videos via
  hybrid embedding aggregation transformer.
\newblock In {\em Medical Image Computing and Computer Assisted
  Intervention--MICCAI 2021: 24th International Conference, Strasbourg, France,
  September 27--October 1, 2021, Proceedings, Part IV 24}, pages 593--603.
  Springer, 2021.

\bibitem{gimeno2023artificial}
Antonio~Z Gimeno-Garc{\'\i}a, Anjara Hern{\'a}ndez-P{\'e}rez, David
  Nicol{\'a}s-P{\'e}rez, and Manuel Hern{\'a}ndez-Guerra.
\newblock Artificial intelligence applied to colonoscopy: Is it time to take a
  step forward?
\newblock {\em Cancers}, 15(8):2193, 2023.

\bibitem{he2016deep}
Kaiming He, Xiangyu Zhang, Shaoqing Ren, and Jian Sun.
\newblock Deep residual learning for image recognition.
\newblock In {\em Proceedings of the IEEE conference on computer vision and
  pattern recognition}, pages 770--778, 2016.

\bibitem{hwang2005automatic}
Sae Hwang, JungHwan Oh, JeongKyu Lee, Yu~Cao, Wallapak Tavanapong, Danyu Liu,
  Johnny Wong, and Piet~C De~Groen.
\newblock Automatic measurement of quality metrics for colonoscopy videos.
\newblock In {\em Proceedings of the 13th annual ACM international conference
  on Multimedia}, pages 912--921, 2005.

\bibitem{jin2017sv}
Yueming Jin, Qi~Dou, Hao Chen, Lequan Yu, Jing Qin, Chi-Wing Fu, and Pheng-Ann
  Heng.
\newblock Sv-rcnet: workflow recognition from surgical videos using recurrent
  convolutional network.
\newblock {\em IEEE transactions on medical imaging}, 37(5):1114--1126, 2017.

\bibitem{jin2020multi}
Yueming Jin, Huaxia Li, Qi~Dou, Hao Chen, Jing Qin, Chi-Wing Fu, and Pheng-Ann
  Heng.
\newblock Multi-task recurrent convolutional network with correlation loss for
  surgical video analysis.
\newblock {\em Medical image analysis}, 59:101572, 2020.

\bibitem{katzir2022estimating}
Liran Katzir, Danny Veikherman, Valentin Dashinsky, Roman Goldenberg, Ilan
  Shimshoni, Nadav Rabani, Regev Cohen, Ori Kelner, Ehud Rivlin, and Daniel
  Freedman.
\newblock Estimating withdrawal time in colonoscopies.
\newblock In {\em European Conference on Computer Vision}, pages 495--512.
  Springer, 2022.

\bibitem{kelner2023semantic}
Ori Kelner, Or~Weinstein, Ehud Rivlin, and Roman Goldenberg.
\newblock Semantic parsing of colonoscopy videos with multi-label temporal
  networks.
\newblock In {\em Proceedings of the IEEE/CVF International Conference on
  Computer Vision}, pages 2599--2606, 2023.

\bibitem{lea2017temporal}
Colin Lea, Michael~D Flynn, Rene Vidal, Austin Reiter, and Gregory~D Hager.
\newblock Temporal convolutional networks for action segmentation and
  detection.
\newblock In {\em proceedings of the IEEE Conference on Computer Vision and
  Pattern Recognition}, pages 156--165, 2017.

\bibitem{lei2018temporal}
Peng Lei and Sinisa Todorovic.
\newblock Temporal deformable residual networks for action segmentation in
  videos.
\newblock In {\em Proceedings of the IEEE conference on computer vision and
  pattern recognition}, pages 6742--6751, 2018.

\bibitem{li2020ms}
Shi-Jie Li, Yazan AbuFarha, Yun Liu, Ming-Ming Cheng, and Juergen Gall.
\newblock Ms-tcn++: Multi-stage temporal convolutional network for action
  segmentation.
\newblock {\em IEEE transactions on pattern analysis and machine intelligence},
  2020.

\bibitem{lin2017focal}
Tsung-Yi Lin, Priya Goyal, Ross Girshick, Kaiming He, and Piotr Doll{\'a}r.
\newblock Focal loss for dense object detection.
\newblock In {\em Proceedings of the IEEE international conference on computer
  vision}, pages 2980--2988, 2017.

\bibitem{liu2010automated}
Xuemin Liu, Wallapak Tavanapong, Johnny Wong, JungHwan Oh, and Piet~C De~Groen.
\newblock Automated measurement of quality of mucosa inspection for
  colonoscopy.
\newblock {\em Procedia Computer Science}, 1(1):951--960, 2010.

\bibitem{ramesh2021multi}
Sanat Ramesh, Diego Dall’Alba, Cristians Gonzalez, Tong Yu, Pietro Mascagni,
  Didier Mutter, Jacques Marescaux, Paolo Fiorini, and Nicolas Padoy.
\newblock Multi-task temporal convolutional networks for joint recognition of
  surgical phases and steps in gastric bypass procedures.
\newblock {\em International journal of computer assisted radiology and
  surgery}, 16:1111--1119, 2021.

\bibitem{rex2023key}
Douglas~K Rex.
\newblock Key quality indicators in colonoscopy.
\newblock {\em Gastroenterology Report}, 11:goad009, 2023.

\bibitem{shine2020quality}
Rebecca Shine, Andrew Bui, and Adele Burgess.
\newblock Quality indicators in colonoscopy: an evolving paradigm.
\newblock {\em ANZ journal of surgery}, 90(3):215--221, 2020.

\bibitem{spadaccini_computer-aided_2021}
Marco Spadaccini, Andrea Iannone, Roberta Maselli, Matteo Badalamenti, Madhav
  Desai, Viveksandeep~Thoguluva Chandrasekar, Harsh~K. Patel, Alessandro
  Fugazza, Gaia Pellegatta, Piera~Alessia Galtieri, Gianluca Lollo, Silvia
  Carrara, Andrea Anderloni, Douglas~K. Rex, Victor Savevski, Michael~B.
  Wallace, Pradeep Bhandari, Thomas Roesch, Ian~M. Gralnek, Prateek Sharma,
  Cesare Hassan, and Alessandro Repici.
\newblock Computer-aided detection versus advanced imaging for detection of
  colorectal neoplasia: a systematic review and network meta-analysis.
\newblock {\em The Lancet. Gastroenterology \& Hepatology}, 6(10):793--802,
  October 2021.

\bibitem{tavanapong2022artificial}
Wallapak Tavanapong, JungHwan Oh, Michael~A Riegler, Mohammed Khaleel, Bhuvan
  Mittal, and Piet~C De~Groen.
\newblock Artificial intelligence for colonoscopy: Past, present, and future.
\newblock {\em IEEE journal of biomedical and health informatics},
  26(8):3950--3965, 2022.

\bibitem{yi2021asformer}
Fangqiu Yi, Hongyu Wen, and Tingting Jiang.
\newblock Asformer: Transformer for action segmentation.
\newblock {\em arXiv preprint arXiv:2110.08568}, 2021.

\bibitem{zhou2021multi}
Wei Zhou, Liwen Yao, Huiling Wu, Biqing Zheng, Shan Hu, Lihui Zhang, Xun Li,
  Chunping He, Zhengqiang Wang, Yanxia Li, et~al.
\newblock Multi-step validation of a deep learning-based system for the
  quantification of bowel preparation: a prospective, observational study.
\newblock {\em The Lancet Digital Health}, 3(11):e697--e706, 2021.

\bibitem{zisimopoulos2018deepphase}
Odysseas Zisimopoulos, Evangello Flouty, Imanol Luengo, Petros Giataganas, Jean
  Nehme, Andre Chow, and Danail Stoyanov.
\newblock Deepphase: surgical phase recognition in cataracts videos.
\newblock In {\em Medical Image Computing and Computer Assisted
  Intervention--MICCAI 2018: 21st International Conference, Granada, Spain,
  September 16-20, 2018, Proceedings, Part IV 11}, pages 265--272. Springer,
  2018.

\bibitem{zorzi2023adenoma}
Manuel Zorzi, Giulio Antonelli, Claudio Barbiellini~Amidei, Jessica Battagello,
  Bastianello German{\`a}, Flavio Valiante, Stefano Benvenuti, Alberto
  Tringali, Francesco Bortoluzzi, Erica Cervellin, et~al.
\newblock Adenoma detection rate and colorectal cancer risk in fecal
  immunochemical test screening programs: An observational cohort study.
\newblock {\em Annals of Internal Medicine}, 176(3):303--310, 2023.

\end{thebibliography}
